%% file: main.tex
\documentclass[11pt]{article}
\usepackage{sustyle}
\usepackage{fullpage}
\usepackage{amsmath}
\usepackage{graphicx}
\usepackage{enumerate}
\usepackage[round]{natbib}
\usepackage{url}  
\usepackage[svgnames]{xcolor}
\usepackage{tcolorbox}
\usepackage{multirow}
\usepackage{microtype}
\usepackage{amsmath,amsfonts,bbm,bbding,bm}
\usepackage{amsthm}
\usepackage{etoolbox}
\usepackage{subfigure}
\usepackage{xcolor}
\usepackage{booktabs} 
\usepackage{nicefrac}   
\usepackage{color} 
\usepackage{enumitem}
\usepackage{dsfont}
\usepackage[linktocpage]{hyperref}

\definecolor{blue}{rgb}{0, 0, 1}
\newtheorem{hypo}{Working Hypothesis}[section]

\renewcommand\d{\mathrm{d}}
  \hypersetup{
  colorlinks,
  citecolor=gray!50!black,
  linkcolor=blue,
  urlcolor=black}
\begin{document}

  \title{On the Algorithmic Bias of Aligning Large Language Models with RLHF: Preference Collapse and Matching Regularization}
    \author{Jiancong Xiao\thanks{University of Pennsylvania.} \and Ziniu Li\thanks{The Chinese University of Hong Kong, Shenzhen.} \and Xingyu Xie\thanks{National University of Singapore.} \and Emily Getzen$^*$ \and Cong Fang\thanks{Peking University.} \and Qi Long\textsuperscript{*}\thanks{Joint corresponding authors. Emails: \texttt{qlong@upenn.edu, suw@wharton.upenn.edu}.} \and Weijie J.~Su\textsuperscript{*\textparagraph}}
    \date{}
  \maketitle
\begin{abstract}

Accurately aligning large language models (LLMs) with human preferences is crucial for informing fair, economically sound, and statistically efficient decision-making processes. However, we argue that the predominant approach for aligning LLMs with human preferences through a reward model---reinforcement learning from human feedback (RLHF)---suffers from an inherent algorithmic bias due to its Kullback--Leibler-based regularization in optimization. In extreme cases, this bias could lead to a phenomenon we term preference collapse, where minority preferences are virtually disregarded. To mitigate this algorithmic bias, we introduce preference matching (PM) RLHF, a novel approach that \textit{provably} aligns LLMs with the preference distribution of the reward model under the Bradley--Terry--Luce/Plackett--Luce model. Central to our approach is a PM regularizer that takes the form of the negative logarithm of the LLM's policy probability distribution over responses, which helps the LLM balance response diversification and reward maximization. Notably, we obtain this regularizer by solving an ordinary differential equation that is necessary for the PM property. For practical implementation, we introduce a conditional variant of PM RLHF that is tailored to natural language generation. Finally, we empirically validate the effectiveness of conditional PM RLHF through experiments on the OPT and Llama‑family models, demonstrating a 29\% to 41\% improvement in alignment with human preferences, as measured by a certain metric, compared to standard RLHF.
\end{abstract}

\begingroup
\renewcommand\thefootnote{}\footnotemark\footnotetext{This paper has been accepted for publication in the \textit{Journal of the American Statistical Association}.
}
\endgroup
\tableofcontents
\input{sec1_intro_jasa}
\input{sec2_preli_jasa}
\input{sec3_pm_jasa}
\input{sec3b_pm_jasa}

\input{sec5_exp_jasa}
\input{sec6_discuss_jasa}

{\small
\section*{Acknowledgments} We thank the Associate Editor and the reviewers for their valuable comments and suggestions. This work was supported in part by NIH grants, R01-EB036016, RF1AG063481 and U01CA274576, NSF DMS-2310679, a Meta Faculty Research Award, and Wharton AI for Business. The content is solely the responsibility of the authors and does not necessarily represent the official views of the NIH.

\bibliographystyle{abbrvnat}
\bibliography{Bibliography-MM-MC}
}

\clearpage
\appendix
\input{appendix}

\end{document}

%% file: sec1_intro_jasa.tex
\section{Introduction}
\label{sec:intro}

Large language models (LLMs) such as ChatGPT-4o \citep{openai2023gpt4} and Claude-3.7 Sonnet \citep{anthropic2024claude} have demonstrated impressive capabilities in tasks such as code generation, basic data analysis, elementary math problem solving, and reasoning~\citep{bubeck2023sparks,chowdhery2023palm,touvron2023llama,bubeck2023sparks,team2023gemini}. The widespread adoption of LLMs for tasks previously considered unlikely to be performed by machines reflects these advanced capabilities. For instance, a recent study revealed that a substantial proportion of machine learning conference reviews had been written or significantly modified by LLMs \citep{liang2024monitoring}.

As a result, LLMs are increasingly influencing decision-making processes across various domains \citep{eloundou2023gpts,duetting2023mechanism}. It is, therefore, crucial to ensure that LLMs accurately reflect human preferences to promote fairness, mitigate the risk of biased outputs, and facilitate economically sound decisions~\citep{arrow2012social}. Human preferences are inherently diverse due to factors such as cultural background, personal experiences, and contexts. It is natural for people from different backgrounds to exhibit varied preferences when making comparisons. For example, \citet{ouyang2022training} and \citet{bai2022training} reported annotator--annotator and annotator--researcher disagreement rates of 37\% and 23\%, respectively, even for trained human labelers. Crucially, the relative proportions of these preferences play a critical role. For example, a restaurant chain deciding between adding a vegan dish or a spicy meat-based option to its menu might base its decision on whether customers have a 90\% to 10\% or a 20\% to 80\% preference for the two options. An LLM that fails to match human preferences could result in biased viewpoints on the tasks assigned to it. Consequently, the suggestions or decisions made by the LLM could suffer from fairness issues or, in this case, lead to undesirable economic outcomes.

Indeed, LLMs have been observed to exhibit significant biases when expressing preferences~\citep{casper2023open,santurkar2023whose, azar2024general,chakraborty2024maxmin}. \citet{casper2023open} noted that LLMs tend to disproportionately favor more frequent items over less frequent ones. This finding is in agreement with the observation by \citet{santurkar2023whose} that opinion distributions generated by LLMs are often highly skewed towards the dominant viewpoints, often assigning over 99\% probability to the dominant opinion and thus failing to capture the diversity of opinions within the population.

In this paper, we aim to understand the origins of bias in LLMs from an algorithmic viewpoint and, more importantly, to develop a method for accurately aligning LLMs with human preferences with provable guarantees. To this end, it is instructive to examine reinforcement learning from human feedback (RLHF)~\citep{ouyang2022training,dong2024rlhf}, the go-to solution for unleashing the power of LLMs by significantly simplifying the process of prompting for general tasks. To put this in perspective, LLMs had existed for at least five years---for example, GPT-3 was introduced in 2020---before the worldwide excitement surrounding the introduction of ChatGPT-3.5, which was enabled by applying standard RLHF to GPT-3. At its core, RLHF trains a reward model using the Bradley--Terry--Luce model \citep{bradley1952rankanalysis,luce2012individual} to learn human preferences based on data from human labelers, and then fine-tunes the pretrained LLM using this reward model.

First, we reveal an inherent bias in standard Kullback--Leibler (KL) divergence-based RLHF, which involves a pretrained LLM as a reference model for computing the divergence. We demonstrate that the reference model, arguably biased since it has not been aligned, can pass its bias to RLHF through the KL-based regularization. In certain cases, we theoretically show that this algorithmic bias can asymptotically amplify to extreme preference imbalances, such as 0\% versus 100\%, a phenomenon we term preference collapse. When preference collapse occurs, the LLM completely disregards minority opinions, which would lead to severe consequences in making fair and economical decisions. Furthermore, we show that merely replacing the KL divergence with a general $f$-divergence \citep{renyi1961measures} does not eliminate this algorithmic bias.

Crucially, this algorithmic bias persists even when the reward model is an \textit{oracle}, meaning it fully represents human preferences. This contrasts with prior research, which often attributed the empirically observed biases in LLMs to the insufficiency of using a single reward model to express complex human preferences \citep{chakraborty2024maxmin,zhong2024provable} or overfitting issues in reward model training, partly due to limited labeling data~\citep{song2023reward,schulman2023icml}.

To address the algorithmic bias and its extreme form, preference collapse, we introduce \textit{preference matching RLHF} as the main contribution of this paper. This novel RLHF method provides a principled approach to precisely aligning LLMs with the reward model's preferences, which comes with provable statistical guarantees. We start by considering a general function that regularizes the policy network, i.e., the probability distribution of responses given a prompt. By deriving a necessary condition for the regularizer to achieve preference matching in the form of an ordinary differential equation and solving it, we show that the regularization must equal the negative logarithm of the response probability distribution, up to certain less relevant terms. We further prove that this form of regularization is sufficient to match the reward model's preferences. Taking the expectation, this regularization term becomes the Shannon entropy. Intuitively, in maximizing the reward plus the preference matching regularization, it seeks to strike a balance between reward maximization and diversifying the generation of responses. Moreover, we extend the differential equation-based strategy to obtain preference matching regularization for a class of generalized Bradley--Terry--Luce/Plackett--Luce models.

In comparison to our approach, a straightforward approach to addressing algorithmic biases is the early stopping of the fine-tuning process in standard RLHF, which appears to be one of OpenAI's approaches based on publicly available information \citep{schulman2023icml}. However, early stopping cannot entirely eliminate the bias in alignment since the ``target'' of standard RLHF is inherently biased, whereas our preference matching RLHF provably sets its global optimization solution to be unbiased.

To examine the empirical performance of our preference matching RLHF, we introduce a variant of PM RLHF by conditioning PM RLHF on the set of responses that resemble natural language. We apply this conditional PM RLHF to fine-tune the OPT \citep{zhang2022opt} and Llama-family \citep{touvron2023llama} models. In our experiments, we use proximal policy optimization (PPO) \citep{schulman2017proximal} to train the model with our RLHF and the standard RLHF. To measure how well an RLHF approach does in terms of preference matching, we introduce a new measure called the preference matching divergence, defined as the KL divergence between the aligned model and the reward model. Using the standard RLHF approach, the preference matching divergence is 2.23 and 1.16 when fine-tuning Llama-2-7B and OPT-1.3B, respectively. Using our preference matching RLHF, the values of this divergence decrease to 1.57 and 0.68, marking improvements of approximately 29\% and 41\%, respectively.

The remainder of the paper is structured as follows. Section \ref{Sec:pre} introduces the basics of LLMs and how standard RLHF operates on them. Section~\ref{sec:pref_policy} introduces the concept of preference matching and obtains preference matching regularization by solving an ordinary differential equation. Section~\ref{sec:subopt} extends preference matching RLHF to language domains, where we resolve a numerical challenge by introducing a conditional variant of our RLHF that is tailored to natural text generation. Experimental results are presented in Section~\ref{sec:experiment}. We conclude the paper with a discussion in Section~\ref{sec:discuss}.

\subsection{Related Work}
\label{sec:related}
\paragraph{Perspective on Bias in Preference Modeling.} The observed bias in RLHF may originate from the assumption that complex human preferences can be captured by a single reward function. \citet{chakraborty2024maxmin} challenged this assumption by providing a reward mismatch lower bound, arguing that diverse human preferences cannot be modeled by a single reward function. Similarly, \citet{zhong2024provable} showed that a single reward function fails to balance the preferences of multiple heterogeneous individuals. \citet{xiao2025theoretical} examined the inconsistency between reward learning and the principles of social choice theory. To address this limitation, \citet{wang2024arithmetic} proposed directional preference alignment with multi-objective rewards to train LLMs for diverse user preferences. \citet{xu2024rlhf} showed that methods such as RLHF can fail to yield desirable outcomes due to violations of the independence of irrelevant alternatives. \citet{liu2025statistical} and \citet{shi2025fundamental} discussed the statistical impossibility of aligning with human preferences.

\paragraph{Diversity in Human Preferences.} Existing alignment approaches predominantly consider the average preference of human annotators, overlooking the rich diversity inherent in human preferences \citep{casper2023open, rlhf_survey2, li2025preserving}. \citet{chakraborty2024maxmin} explored the factors contributing to this diversity, which often arises from various social and cultural backgrounds \citep{aroyo2023dices}. These factors include socio-demographic backgrounds, personal bias and context subjectivity, imperfect preferences, and linguistic ambiguity and missing context. Additional information on these factors is provided in the online Appendix.

%% file: sec2_preli_jasa.tex
\section{Preliminaries}
\label{Sec:pre}
LLMs are large-scale deep learning architectures trained on extensive datasets \citep{vaswani2017attention,kenton2019bert}. They are built upon the Transformer architecture \citep{vaswani2017attention}, which is designed to capture the context and relationships within textual sequences. This enables the models to extract and comprehend the meanings of words and phrases effectively.

The generation of sentences by LLMs relies on predicting the next token. Tokenization \citep{grefenstette1999tokenization} is the process of converting text into smaller units called tokens. Tokens can be words, subwords, characters, or other meaningful units. This process is a crucial step in natural language processing (NLP) and language modeling because it transforms raw text data into a format that can be processed by machine learning algorithms and models. For instance, the sentence ``I just got a funky phone case!'' can be tokenized into [``I'', ``\_just'', ``\_got'', ``\_a'', ``\_fun'', ``ky'', ``\_phone'', ``\_case'', ``!''], where ``\_'' represents a space. Formally, tokenization can be expressed as $y=[T_1, T_2, T_3, \ldots, T_L]$, where $L$ is the length of the sentence. Utilizing the sequence of preceding tokens, an LLM predicts the subsequent token. For example, the LLM assigns a probability to $p(T_3 \mid \text{``I'', ``\_just''})$ across a range of possible next tokens such as [``\_got'', ``\_took'', $\cdots$] from its vocabulary.

\subsection{Reinforcement Learning from Human Feedback}
\label{sec:rlhf}
RLHF is a promising approach to align LLMs with human preferences, ensuring that the models are useful and safe for their human users \citep{ouyang2022training}. The pre-training fine-tuning framework for learning LLMs typically consists of the following three steps: (1) supervised fine-tuning; (2) preference and reward modeling; and (3) policy learning and RLHF fine-tuning.

\paragraph{Step 1: Supervised Fine-Tuning (SFT).} RLHF typically begins with a generic pre-trained language model, which is fine-tuned using supervised learning on a high-quality dataset to obtain a model $\pi_{\textnormal{SFT}}$.

\paragraph{Step 2: Preference and Reward Modeling.} Let $x$ be a prompt given to an LLM and $y$ be the output response. In the second step, the SFT model is prompted with prompts $x$ to produce pairs of answers $(y_1, y_2) \sim \pi_{\textnormal{SFT}}$. For instance, consider the prompt $x$ being the question: ``Please suggest a drink.'' Two possible responses to this prompt are $y_1$ = ``I would suggest coffee'' and $y_2$ = ``I would suggest tea,'' respectively. After that, human labelers express their preferences for one of the two answers, labeled as $y_\textnormal{w} \succ y_\textnormal{l}$, where $y_\textnormal{w}$ and $y_\textnormal{l}$ denote the winner and loser among $(y_1, y_2)$, respectively.

The preferences are assumed to be generated by some true reward model $r(x, y)$, which we do not have access to. In standard RLHF \citep{ouyang2022training}, the preference model is assumed to be the Bradley--Terry--Luce (BTL) model \citep{bradley1952rankanalysis,luce2012individual}, which assumes that the human preference distribution can be written as
\begin{equation}
\label{eq:BT}
\mathbb{P}(y_1 \succ y_2|y_1, y_2,x)=\frac{\exp(r(x,y_1))}{\exp(r(x,y_1))+\exp(r(x,y_2))}.
\end{equation}
\noindent By collecting an offline dataset of comparisons $D_\textnormal{comp} = \{x^i , y_\textnormal{w}^i , y_\textnormal{l}^i\}_{i=1}^n$, we
can parameterize a reward model $r_\theta(x,y)$ and estimate the parameters via maximum likelihood. The problem is formulated as a binary classification problem. The negative log-likelihood function is
\begin{equation}
\label{eq:rewardloss}
    \min_\theta - \mathbb{E}_{(x,y_\textnormal{w}, y_\textnormal{l})\sim D_\textnormal{comp}}[\log(\sigma(r_\theta(x,y_\textnormal{w})-r_\theta(x,y_\textnormal{l})))],
\end{equation}
where $\sigma(u) := 1/(1+ \exp(-u))$ is the logistic function. Hence, we omit the dependence of the reward function $r_{\theta}$ on $\theta$ because this paper does not discuss how the reward model is trained. 

\paragraph{Step 3: Policy Learning and RLHF Fine-tuning.} Let $\pi_{\phi}(y|x)$ be the probability distribution of the responses given a prompt $x$, where $\phi$ denotes the weights of the LLM. The goal of (unregularized) RLHF fine-tuning is to maximize the expected reward, i.e., $\max_\phi \mathbb{E}_{y\sim\pi_{\phi}(\cdot|x)} r(x,y)$. To mitigate over-optimization of the reward model, additional regularization should be added to the objective function, i.e., $\max_\phi \mathbb{E}_{y\sim\pi_{\phi}(\cdot|x)} r(x,y)-\text{Regularizer},$ where, unless specified otherwise, the expectation is taken over the randomness of $x$ and, conditional on $x$, the randomness of $y$ sampled from $\pi_{\phi}(\cdot|x)$. In practice, $x$ can be either sampled from a fixed database of prompts \citep{rafailov2023direct} or adaptively based on prior responses \citep{xiong2023iterative}. RLHF uses a KL penalty between the RLHF model and the reference model, which is the pretrained or SFT model, at each token as the regularizer. The loss function of regularized RLHF fine-tuning is
\begin{equation}\label{eq:prlloss}
\max_\phi \mathbb{E}_{y\sim\pi_{\phi}(\cdot|x)} r(x,y)-\beta D_{\text{KL}}(\pi_\phi (y|x)\| \pi_{\textnormal{ref}}(y|x)),
\end{equation}
where $\beta > 0$ is a parameter controlling the deviation from the base reference policy $\pi_{\textnormal{ref}}$, namely the initial SFT model $\pi_{\text{SFT}}$. Above, the Kullback--Leibler divergence between $\pi_\phi (y|x)$ and $\pi_{\textnormal{ref}}(y|x)$ is defined to be $\mathbb{E}_{y\sim\pi_{\phi}(\cdot|x)}\big[\log (\pi_\phi (y|x))-\log (\pi_{\textnormal{ref}} (y|x))\big]$.
In practice, $\pi_\phi (y|x)$ is calculated token-wise, let $y=[T_1,\cdots,T_L]$, it can be expressed in
\begin{equation*}
\log (\pi_\phi (y|x))=\sum_{i=1}^L\bigg[ \log\pi_\phi(T_i\mid T_{i-1},\cdots,T_1,T_0)\bigg],
\end{equation*}
where we further denote $T_0=x$. The same notation is used for $\pi_{\textnormal{ref}}(y|x)$. For clarity, the term RLHF problem specifically refers to the problem defined in \eqref{eq:prlloss}. When comparing various RLHF approaches, \eqref{eq:prlloss} is referred to as either standard RLHF or KL RLHF.

%% file: sec3_pm_jasa.tex
\section{Preference Matching RLHF}
\label{sec:pref_policy}

We start by presenting a motivating example. For a question $x$, formally referred to as a prompt, let $y_1$ and $y_2$ be two possible responses. Suppose 60\% of human labelers prefer $y_1$ over $y_2$ for this binary comparison. To ensure fairness and protect minority preferences, it would be desirable for the model's output probabilities to reflect this distribution (for example, $\pi(y_1|x)/\pi(y_2|x)=60\%/40\%$). We now formally define the following condition.
\begin{definition}[Preference Matching]
\label{def:preference_matching}
A policy $\pi(y|x)$ is said to be preference matching (PM) if, for any prompt $x$ and any pair of responses $y_1, y_2$, it satisfies
\begin{equation*}
\frac{\pi(y_1|x)}{\pi(y_1|x)+\pi(y_2|x)}=\mathbb{P}(y_1 \succ y_2|x,y_1,y_2),
\end{equation*}
where $\mathbb{P}(y_1 \succ y_2|x,y_1,y_2)$ denotes the target human preference distribution.
\end{definition}

\begin{remark}
From a practical perspective, we only require Definition~\ref{def:preference_matching} to hold for prompts that do not admit a single correct response, with the response distribution depending on personal opinions or preferences. Another reason for not requiring it to hold for all prompts is that the reward model might not generalize well for some prompts.
\end{remark}
For general human preferences, a preference matching policy as defined in Definition~\ref{def:preference_matching} may not always exist due to conflicts across different pairwise comparisons. To further explore which human preferences can be matched by LLMs,  we first introduce the extension of the BTL model to multi-comparison settings, which yields the Plackett--Luce (PL) model \citep{plackett1975analysis,luce2012individual}. The PL model outputs the preference in the form of a ranking of different items, which will not be used in our paper. Therefore, we provide the standard assumption of the PL model in the appendix. For simplicity, we consider an equivalent formulation of the PL model that assumes the preference of one response over others: $
\mathbb{P}(y|x)=\exp(r(x,y))/\sum_{y'} \exp(r(x,y')),$
where the sum is over all possible responses to the prompt $x$.

Given a human preference, if a preference-matching policy exists, it can be represented by the PL model. Conversely, if human preferences follow the PL model assumption, a preference-matching policy necessarily exists. Therefore, throughout the remainder of the paper, we assume that human preferences follow the PL model. We refer to the policy $\pi^\star(y \mid x) = \exp(r(x, y))/\sum_{y'} \exp(r(x, y'))$
as the \emph{preference-matching policy}.

Given a reward model, consider the simple scenario of optimizing the expected reward without any regularization. Assuming that the reward model accurately reflects this 60\% versus 40\% preference between the two responses, it will assign a slightly higher reward $r(x, y_1)$ to the first response than $r(x, y_2)$ to the second one. Maximizing the reward without any regularization would lead the LLM to exclusively prefer the majority opinion and completely disregard any minority opinions.

\begin{proposition}\label{prop:reward100}
Let $\phi^\star$ be an optimal solution to the unregularized reward maximization problem $\max_{\phi}\mathbb{E}_{y\sim\pi_\phi(\cdot|x)} r(x,y)$, where the expectation is over the randomness of both $x$ and $y$ following the conditional distribution $\pi_\phi(\cdot|x)$. For a fixed $x$, with probability one, $\pi_{\phi^\star}(y|x)$ outputs some response $y^\star$ with the highest reward $r(x, y^\star)$. 
\end{proposition}

Here, we assume sufficient expressivity of the LLMs and that the expected reward is maximized exactly, as detailed in Hypotheses~\ref{hy:flex} and \ref{hy:opt} in Section~\ref{sec:derive}. Although this result is almost trivial, it implies that regularization is necessary to match the preferences of the reward model. Formally, we consider adding a regularizer $R(\pi_\phi(y|x))$ to the reward maximization problem:
\begin{equation}\label{eq:addregular}
\max_\phi \mathbb{E}_{ y\sim\pi_{\phi}(\cdot|x)} \big[r(x,y)+R(\pi_\phi (y|x))\big].
\end{equation}
Specifically, our goal is to determine the form of $R(\cdot)$ such that the global solution to \eqref{eq:addregular} is statistically unbiased. We seek to identify what regularization structure would ensure that the solution possesses certain desirable properties---for example, satisfying the PM condition. As such, our investigation is entirely statistically motivated.

Following Definition~\ref{def:preference_matching}, we call $R$ a PM regularizer if solving \eqref{eq:addregular} leads to a PM policy $\pi_{\phi^\star}(y|x)$. Assuming that the reward model faithfully reflects the preferences learned from human-labeled data, PM regularization ensures that the LLM fine-tuned by an RL algorithm maximizing \eqref{eq:addregular} would have (approximately) the same preferences as the human labelers.

\subsection{Preference Matching Differential Equation}
\label{sec:derive}

Now, we aim to find all possible PM regularizers by finding a necessary condition for the PM property. To this end, we first propose two hypotheses to facilitate our discussion.

\begin{hypo}[Sufficiency of expressivity]\label{hy:flex}
For any $x$, we assume that the policy of the LLM, $\pi_{\phi}(y|x)$, can represent an arbitrary probability distribution function over the universe of responses by varying the parameter $\phi$.
\end{hypo}

\begin{hypo}[Soundness of optimization]\label{hy:opt}
The fine-tuning process of the LLM can find the global solution to the optimization problem \eqref{eq:addregular}. 
\end{hypo}

These two working hypotheses allow us to obtain an almost closed-form expression of the optimal policy for \eqref{eq:addregular}. Under Hypothesis \ref{hy:flex}, any policy $\phi$ is feasible in the sense that there exists $\phi$ such that $\pi = \pi_\phi$. This suggests that we can directly optimize the policy $\pi$ instead of $\phi$ that parametrizes it. Hence, we consider the following program that is more amenable to analysis than \eqref{eq:addregular}:
\begin{equation}\label{eq:rh_p}
\max_{\pi} \mathbb{E}_{y\sim \pi(\cdot|x)} \bigg[ r(x,y)+R(\pi(y|x)) \bigg],
\end{equation}
where we use $\pi$ rather than $\pi_\phi$ to highlight that $\pi$ itself becomes the decision variable in this optimization program.

Next, we study under what conditions the exact solution to \eqref{eq:rh_p} would satisfy the PM property. We can write \eqref{eq:rh_p} in the form of a conditional expectation:
\[
\mathbb{E}_{y\sim \pi(\cdot|x)} \bigg[ r(x,y)+R(\pi(y|x)) \bigg] =  \E_x \E_{y\sim \pi(\cdot|x)} \bigg[ r(x,y)+R(\pi(y|x)) \Big{|} x\bigg],
\]
where the second $\E_{y\sim \pi(\cdot|x)}$ is interpreted as the expectation over only the randomness of $y$. Thus, it would be sufficient to find the optimal $R$ that maximizes 
\begin{equation}\label{eq:fixed_x_eq}
\E_{y\sim \pi(\cdot|x)} \bigg[ r(x,y)+R(\pi(y|x)) \Big{|} x\bigg]
\end{equation}
for each $x$. For a fixed $x$, without loss of generality, we assume the distribution $\pi(\cdot|x)$ is a multinomial distribution on $y_1, \ldots, y_k$ with probabilities $\pi_1, \ldots, \pi_k$, respectively. 
By Hypothesis~\ref{hy:opt}, \eqref{eq:fixed_x_eq} can be exactly minimized over such multinomial distributions. Denoting by $r_i = r(x, y_i)$, therefore \eqref{eq:fixed_x_eq} is equivalent to
\begin{equation}\label{eq:disc_pik}
\max_{\pi_1, \ldots, \pi_k} \sum_{i=1}^k \pi_i(r_i + R(\pi_i)),
\end{equation}
subject to the constraints $\sum_{i=1}^k \pi_i = 1$ and $\pi_i \ge 0$.

Assuming that $R$ is sufficiently differentiable, we aim to find the necessary condition on $R$ that leads to PM regularization based on \eqref{eq:disc_pik}. Consider the Lagrangian of \eqref{eq:disc_pik}: $\textstyle L = \sum_{i=1}^k \pi_i(r_i + R(\pi_i)) + \lambda \left( \sum_{i=1}^k \pi_i - 1 \right)$. Here, we omit the constraints since they will be satisfied automatically. The first-order conditions $\partial L/\partial \pi_i = r_i + R(\pi_i) + \pi_iR'(\pi_i) + \lambda = 0$
must be satisfied by the PM policy $\pi^\star$. Thus, we get
\begin{equation}\label{eq:hr}
R(\pi_i^\star) + \pi_i^\star R'(\pi_i^\star) = -r_i -\lambda
\end{equation}
for all $i$, where $\pi^\star_i := \pi^\star(y_i|x) = \exp(r_i)/\sum_{j=1}^k \exp(r_j)$. Conversely, we can write 
$$\textstyle r_i = \log\left(\pi^\star_i\sum_{j=1}^k \exp(r_j) \right) = \log \pi_i^\star + c,$$
where $c = \log (\sum_{j=1}^k \exp(r_j))$ does not depend on $i = 1, \ldots, k$. This expression of $r_i$ together with \eqref{eq:hr} gives
\begin{equation}\label{eq:c_ind_al}
R(\pi_i^\star) + \pi_i^\star R'(\pi_i^\star) = -\log \pi_i^\star + c',
\end{equation}
where $c' = -c -\lambda$ does not depend on $i$. 

Taking $\pi_i^\star$ as a variable, as it can take any value in $[0,1]$ in principle, and differentiating both sides of \eqref{eq:c_ind_al}, we show that $R$ must satisfy the following second-order ordinary differential equation:
\begin{equation*}\label{eq:pm}
\pi  R''(\pi) + 2R'(\pi) + \frac{1}{\pi} = 0,
\end{equation*}
which we call the PM differential equation.

\subsection{Preference Matching Regularization}
\label{sec:reg_exps}

To solve the PM differential equation, note that $\pi R'' + 2R' = (\pi R)''$. Hence, we get
\begin{equation}\label{eq:r_pi_exp}
R(\pi) = \frac{1}{\pi} \iint\pi R'' + 2R'\d\pi = \frac{1}{\pi}  \iint - \frac{1}{\pi} \d\pi = -\log\pi +C_1+ \frac{C_2}{\pi},
\end{equation}
where $C_1$ and $C_2$ are constants. In words, $\pi^\star(y|x) \equiv \exp(r(x,y))/\sum_{y'}\exp(r(x,y'))$ is the global solution to \eqref{eq:fixed_x_eq} only if $R$ takes the form above. 

It is important to recognize that the form of $R$ here does not depend on the specific probabilities $\pi_i^\star$, which implicitly depend on the prompt $x$. This is a convenient fact that facilitates the implementation of the PM regularization. If there were such dependence, it would be very complicated to compute $R$ since it would require knowledge of the values of $\pi_i^\star$ from the reward model.

Notably, the derivation of \eqref{eq:r_pi_exp} is conditional on $x$, which implies that the constants $C_1$ and $C_2$ can depend on $x$. Taken together, our discussion above suggests the following theorem, of which the sufficiency part is proved in the appendix. 

\begin{theorem}\label{thm:penaltyform}
Under Hypotheses \ref{hy:flex} and \ref{hy:opt}, solving \eqref{eq:addregular} yields the PM policy if and only if 
\begin{equation}\label{eq:hc_log}
R(\pi) = -\log\pi +C_{1,x}+ \frac{C_{2,x}}{\pi},
\end{equation}
where $C_{1,x}$ and $C_{2,x}$ are arbitrary constants depending on $x$.
\end{theorem}

\begin{remark}
For the last term in $R(\pi)$, note that 
\begin{equation*}
    \mathbb{E}_{y\sim \pi(\cdot|x)}\frac{C_{2,x}}{\pi(y|x)}=\E_x \left[\sum_{y} \pi(y|x)\frac{C_{2,x}}{\pi(y|x)} \right] = \E_x \left[ k C_{2,x} \right]
\end{equation*}
is a constant that is independent of the decision variable $\pi$. Therefore, this term does not influence the optimization of \eqref{eq:rh_p} at all.
\end{remark}

Taking the expectation over the randomness of $y$ conditional on $x$, the term $-\log \pi$ in the regularization is equal to the Shannon entropy of the policy $\pi(\cdot|x)$. The entropy in the maximization problem would encourage more diverse responses as opposed to the modal collapse in Proposition~\ref{prop:reward100}.

Now we formally define PM RLHF by bringing back the parameter $\phi$ into the maximization problem.
\begin{definition}
We refer to an RLHF variant as PM RLHF under the PL model if it seeks to optimize the following optimization program: 
\begin{equation}\label{eq:pplRL}
    \max_ {\phi} \mathbb{E}_{y\sim \pi_\phi(\cdot|x)}\bigg[r(x,y)-\log(\pi_\phi(y|x))+C_{1,x}+\frac{C_{2,x}}{\pi_\phi(y|x)}\bigg].
\end{equation}
\end{definition}

\paragraph{A Dual Perspective.} Let $C_{1,x}=C_{2,x}=0$, we can interpret the PM RLHF objective \eqref{eq:pplRL} through the lens of robust optimization. Note that the dual of \eqref{eq:pplRL} is
\begin{equation*}
\begin{aligned}
\max_ {\phi} \min_{r'(x,y)} \mathbb{E}_{y \sim \pi_{\phi}(\cdot|x)}r'(x,y)\ \ 
\text{s.t.}\  \text{log-sum-exp}(r(x, \cdot)-r'(x, \cdot))\leq\epsilon
\end{aligned}
\end{equation*}
for some constant $\epsilon$, where the log-sum-exp function equals $\textstyle \log \left[\sum_{y}\exp(r(x,y)-r'(x,y)) \right]$. The constraint can be interpreted as an adversarial perturbation of the rewards for robust optimization.
In the appendix, we obtain the dual formulation and discuss several examples of such objective functions with different constants $C_{1,x}$ and $C_{2,x}$.

\subsection{Extension to Response-Dependent Regularization}
\label{sec:res_dep}

The discussion in Section~\ref{sec:reg_exps} implicitly assumes the dependence of $R$ on the prompt $x$. Now, we extend this dependence to both $x$ and the response $y$, and it would be of interest to find a larger class of PM regularization due to this extended scope:
\begin{equation}\label{eq:addregular_ext}
\max_\phi \mathbb{E}_{ y\sim\pi_{\phi}(\cdot|x)} \big[r(x,y)+R_{x,y}(\pi_\phi (y|x))\big].
\end{equation}
This, in particular, includes the standard KL RLHF \eqref{eq:prlloss}.

Just as with its response-independent counterpart, Hypothesis \ref{hy:flex} allows us to write \eqref{eq:addregular_ext} in the following $\phi$-independent form:
$\max_{\pi} \mathbb{E}_{y\sim \pi(\cdot|x)} \big[ r(x,y)+R_{x,y}(\pi(y|x)) \big]$. Next, Hypothesis~\ref{hy:opt} shows that the $\phi$-independent optimization program above can be further reduced to
\begin{equation}\label{eq:opt_i_in}
\max_{\pi_1, \ldots, \pi_k} \sum_{i=1}^k \pi_i(r_i + R_i(\pi_i)),
\end{equation}
subject to the constraints $\sum_{i=1}^k \pi_i = 1$ and $\pi_i \ge 0$, where $R_i(\cdot) := R_{x, y_i}(\cdot)$.

To admit the PM policy as its solution, just as with \eqref{eq:c_ind_al}, the optimization program \eqref{eq:opt_i_in} shows that $R_{i}$ must obey $R_i(\pi) + \pi R_i'(\pi) +\log\pi = c'$ for all $1 \le i \le k$. Solving this differential equation gives
\begin{equation}\label{eq:r_pi_exp2}
R_i(\pi) = \frac{1}{\pi} \int\pi R_i' + \pi R_i'\mathrm{d}\pi = -\log\pi +C_1+ \frac{C_{2,i}}{\pi},
\end{equation}
where $C_1=c'-1$ is independent of $i$, while $C_{2,i}$ could depend on $i$. As the derivation of \eqref{eq:r_pi_exp2} is conditional on $x$, both the constants $C_1$ and $C_{2,i}$ can depend on $x$. Taken together, our discussion above suggests the following theorem, which we prove in the appendix.

\begin{theorem}\label{thm:penaltyformy}
Under Hypotheses \ref{hy:flex} and \ref{hy:opt}, solving \eqref{eq:addregular_ext} yields the PM policy if and only if
\begin{equation*}
R_{x,y}(\pi) = -\log\pi +C_{1,x}+ \frac{C_{2,x,y}}{\pi},
\end{equation*}
where $C_{1,x}$ is an arbitrary constant depending only on $x$, and $C_{2,x,y}$ is an arbitrary constant depending on $x$ and $y$.
\end{theorem}

Since the KL penalty used in \eqref{eq:prlloss} is included as an instance of \eqref{eq:addregular_ext} but cannot be expressed as in Theorem~\ref{thm:penaltyformy}, we have the following corollary.

\begin{proposition}
The standard KL RLHF \eqref{eq:prlloss} does not yield the PM policy.
\end{proposition}

It is not clear how to leverage the dependence on both the prompt and response for a choice of nonzero $C_{2,x,y}$ with good practical performance. For example, one possibility is to set $C_{2,x,y} = C \pi_{\phi'}(y|x)$ for some fixed parameter $\phi'$ for the LLM. We leave this question for future research.

Finally, we provide a brief comparison with an existing alignment approach designed to handle diverse human preferences: MaxMin RLHF~\citep{chakraborty2024maxmin}. Let the human population be $\mathcal{U}$. Let $u$ represents the human subpopulation defined over a finite discrete set $\mathcal{U} := \{\mathcal{H}_1, \mathcal{H}_2,\ldots , \mathcal{H}_{|\mathcal{U}|}\}$, such
that $\mathcal{H} = \bigcup_{u=1}^{|\mathcal{H}|}\mathcal{H}_u$. The cardinality of the set $\mathcal{U}$ represents
the number of sub-populations in the total human population $\mathcal{H}$. Assume that the sub-population $\mathcal{H}_u$ can be modeled by a BT model with an associated reward function $r_u(x, y)$, for all $u$. Under this assumption, the loss function of MaxMin RLHF can be formulated as: $\max_\pi \min_u \mathbb{E}_x \mathbb{E}_{y \sim \pi(\cdot \mid x)} \left[ r_u(x, y) -\beta \mathbb{D}_\textnormal{KL}\big(\pi(\cdot \mid x)\,\|\,\pi_\textnormal{ref}(\cdot \mid x)\big) \right].$
The main difference between MaxMin RLHF and PM RLHF lies in their fundamental goals. MaxMin RLHF aims to address group-level bias across distinct preference groups, whereas our approach focuses on eliminating individual-level bias within a single preference group.

%% file: sec3b_pm_jasa.tex
\section{Conditional Preference Matching RLHF}
\label{sec:subopt}
When applying PM RLHF in practice, we encounter a numerical issue that is inherent to the nature of text data. A direct application of PM RLHF in our experiments fails to produce human-like text. This numerical issue corresponds to high values of perplexity (see Section \ref{sec:experiment} for the definition of perplexity). For the Llama-2-7B model \citep{touvron2023llama}, the perplexity scores for LLMs trained by PM RLHF and KL RLHF are 7.15 and 5.48, respectively. Further experimental details are provided in the appendix.

A possible cause is that the number of all responses is excessively large. In principle, the number can be up to $|\mathcal{V}|^L$ for responses of length no more than $L$, where $\mathcal{V}$ is the token vocabulary. For example, $|\mathcal{V}| = 50,272$ for the OPT-1.3B model, and $|\mathcal{V}| = 32, 000$ for the Llama series models, and $|\mathcal{V}| = 50, 257$ for GPT-2 and GPT-3.5 series models \citep{openai2023gpt4}. However, most such responses are meaningless and the meaningful and natural text occupies only a negligible fraction. In particular, the reward model might not be generalized well on the meaningless text. Consequently, PM RLHF does not exclude the generation of meaningless text, as the reward model might assign positive probability to such text.

\subsection{Conditional Preference Matching}

To tackle the numerical issue above, we propose to divide the responses space, which we refer to as $\mathcal{Y}$, into $\mathcal{M}(x)$ and $\mathcal{Y}\setminus\mathcal{M}(x)$. The former contains all well-formed natural language responses to prompt $x$, while the latter consists of malformed outputs such as nonsensical text. Although the precise formulation of these sets remains to be determined, for the moment, we assume that the reward model behaves well on $\mathcal{M}(x)$ in the sense that the rewards on this set reflects well the human labelers' preferences.

Our hypothesis is that by \textit{conditioning} PM RLHF on the set where the reward model behaves well, the numerical issue would be resolved. As such, we consider a constrained variant of \eqref{eq:pplRL}:
\begin{equation*}
\begin{aligned}
    \max_{\pi}\  & \mathbb{E}_{y\sim \pi(\cdot|x)} \bigg[r(x,y)-\log(\pi (y|x))+C_{1,x}+\frac{C_{2,x}}{\pi (y|x)}\bigg],\\
    \text{s.t.}\  & \pi(y|x)\geq 0,\ \text{if}\  y\in\mathcal{M}(x),\ \pi(y|x)=0,\ \text{if}\  y\notin\mathcal{M}(x).
\end{aligned}
\end{equation*}
Here, the first constraint is automatically satisfied by parametrization of neural networks. However, this program is difficult to optimize due to a large number of constraints. Instead, we consider an unconstrained relaxation of the above optimization program: 
{\footnotesize\begin{equation}
\label{eq:nontabular}
\begin{aligned}
    \max_{\pi}\  & \mathbb{E}_{y\sim \pi(\cdot|x)} \bigg[r(x,y)-\log (\pi(y|x)) \mathds{1}(y\in\mathcal{M}(x))-\log(\pi(y|x)/\epsilon)\mathds{1}(y\notin\mathcal{M}(x))+C_{1,x}+\frac{C_{2,x}}{\pi(y|x)}\bigg],
\end{aligned}
\end{equation}}
where $\epsilon>0$ is a small number. 

For meaningful responses in $\mathcal{M}(x)$, the regularization $\log (\pi(y|x))$ remains the same as in the original PM RLHF. In contrast, when the response is meaningless, the regularization becomes $\log(\pi(y|x)/\epsilon)$, which penalizes heavily if $\pi(y|x) \ne 0$ as $\epsilon \goto 0$.

The following theorem shows that the global solution to \eqref{eq:nontabular} is preference matching in the conditional sense. Its proof is relegated to the appendix.
\begin{theorem}\label{thm:ym}
Under the same assumptions as in Theorem~\ref{thm:penaltyform}, the optimal solution $\pi^\star(y|x)$ to \eqref{eq:nontabular} conditional on $\mathcal{M}(x)$ satisfies
\begin{equation*} 
\pi^\star(y|\mathcal{M}(x),x)=\frac{\exp(r(x,y))}{\sum_{y'\in\mathcal{M}(x)}\exp(r(x,y'))}
\end{equation*}
for any $y\in\mathcal{M}(x)$.
\end{theorem}
This theorem holds for an arbitrary choice of $\mathcal{M}(x)$, not necessarily that containing only meaningful responses for the prompt $x$. In particular, by letting $\epsilon \goto 0$, $\pi^\star(y|x)$ tend to output responses only in the meaningful set $\mathcal{M}(x)$, that is, $\pi^\star(\mathcal{M}(x)|x) \goto 1$. As such, we have $\pi^\star(y|x)=\exp(r(x,y))/\sum_{y'\in\mathcal{M}(x)}\exp(r(x,y'))$
in the limit $\epsilon \goto 0$.

\subsection{Parameterization of Conditional PM RLHF}
To conclude this section, we discuss some practical aspects of implementing the conditional PM RLHF. Following Example 2 in Section~\ref{sec:reg_exps}, we set $C_{1,x} = C_1 = -\mathbb{E}_{y\sim \pi(\cdot|x)}r(x,y)$, where the expectation is over the randomness of both $x$ and $y$, and $C_{2,x}=0$~\citep{ouyang2022training}. The reward now has mean zero and is still denoted by $r(x, y)$ for simplicity. The objective function becomes
\[
r(x,y)-\log (\pi(y|x)) \mathds{1}(y\in\mathcal{M}(x))-\log(\pi(y|x)/\epsilon)\mathds{1}(y\notin\mathcal{M}(x)).
\]
To implement the conditional PM RLHF above, both the regular set $\mathcal{M}(x)$ and the relaxation parameter $\epsilon$ need to be set. We leverage the reference model to serve as a basis to characterize regular responses. Specifically, we assume that a regular response would yield a relatively large value of $\pi_{\textnormal{ref}}(y|x)$, which suggests setting the regular set as $\mathcal{M}(x) := \{y:\pi_{\textnormal{ref}}(y|x)\geq \alpha\}$ for some $\alpha > 0$. It is worthwhile mentioning that Theorem~\ref{thm:ym} remains true for this choice of regular set.

Next, we discuss how to choose $\epsilon$. First of all, while its choice might influence the likelihood of generating a response, it does not impact the conditional preference matching of those responses such that $\pi_{\textnormal{ref}}(y|x) \ge \alpha$. In general, a very small value of $\epsilon$ is desired from a theoretical perspective since it penalizes the generation of meaningless responses. However, this also increases the burden of numerical optimization. To balance the theoretical guarantee and practical tractability, we let $\epsilon$ depend on both $x$ and $y$ and set $\epsilon(x,y)=\pi_\textnormal{ref}(y|x)$ to serve as a computationally feasible calibration for relaxation. This would ensure that the model remains reasonably close to the reference model, avoiding significant deviations that could impact the model robustness and reliability. Finally, we propose the following conditional PM RLHF that we will use in our experiments:
\begin{definition}
We refer to 
\begin{equation*}\label{eq:consistantRLHF}
\begin{aligned}
    \max_\phi  & \mathbb{E}_{y\sim \pi_\phi(\cdot|x)} \bigg[r(x,y)-\log(\pi_\phi(y|x))\mathds{1}(\pi_{\textnormal{ref}}(y|x)\geq \alpha)-\log\bigg(\frac{\pi_\phi(y|x)}{\pi_{\textnormal{ref}}(y|x)}\bigg)\mathds{1}(\pi_{\textnormal{ref}}(y|x)< \alpha)\bigg].
\end{aligned}
\end{equation*}
as the conditional PM RLHF.
\end{definition}
The conditional PM RLHF has only one explicit tuning parameter, $\alpha$. Fortunately, any choice of this parameter would yield a preference matching policy on responses with probability $\pi_{\textnormal{ref}}(y|x)$ above the threshold. In Section~\ref{sec:experiment}, we experiment with several choices of $\alpha$ and show that the results are robust with respect to the choice of this parameter.

%% file: sec5_exp_jasa.tex
\section{Experiments}\label{sec:experiment}
In this section, we demonstrate the effectiveness of conditional PM RLHF through a comparison with KL RLHF. The experiments were conducted on a computing infrastructure equipped with four A100 GPUs, each with 80GB of memory. Throughout the experimental process, each stage of RLHF follows the standardized configurations established in the DeepSpeed-Chat framework \citep{yao2023deepspeed}.

For experiments involving a publicly available dataset, we employ a random partitioning strategy to divide it into three distinct subsets: 20\% is allocated for SFT, 40\% is allocated to learning the reward functions, and the remaining 40\% is reserved for the reward maximization process.

\paragraph{Reward Model.} The reward model is trained on a dataset of human preferences. In the second stage, the reward model is trained using a network architecture identical to the LLMs, but with a different prediction head that outputs a scalar reward value. The training process involves minimizing the loss function defined in \eqref{eq:rewardloss}, with early stopping applied to prevent over-optimization \citep{schulman2023icml}. It is worth noting that our theory requires $\beta=1$, and our experiments focus on this case. However, note that \eqref{eq:prlloss} is equivalent to
\begin{equation*}
    \max_\phi \mathbb{E}_{y\sim\pi_{\phi}(\cdot|x)} r_\theta(x,y)/\beta- D_{\text{KL}}(\pi_\phi (y|x)\| \pi_{\textnormal{ref}}(y|x)).
\end{equation*}
Ablation studies are also performed by experimenting with values of $\beta \neq 1$. In this setting, $r_\theta(x,y)/\beta$ can be considered a better approximation of the true reward function $r^*(x,y)$ compared to $r_\theta(x,y)$ itself. Consequently, we utilize $r(x,y) = r_\theta(x,y)/\beta$ as the reward function in our experiments, where $\beta$ is treated as a hyperparameter.

Another hyperparameter to tune is the threshold $\alpha$, which can be set to $1 - \texttt{top\_p}$, where \(\texttt{top\_p}\) is the nucleus sampling parameter used in SFT \citep{holtzmancurious}. The intuition for this choice is that it ensures natural and coherent generation by restricting sampling to high-probability tokens, while responses in the \(1 - \texttt{top\_p}\) tail are typically low-quality or unnatural.

\subsection{Evaluation Metrics}
\paragraph{Preference Matching Divergence.} To evaluate the alignment between the reward model's preferences and those of the LLMs, we introduce the preference matching divergence metric. For a given prompt $x$, the procedure involves generating a pair of responses, denoted as $y_1$ and $y_2$. Consistent with the previous notation, the preference of the LLMs is defined as:
 \[p_\text{llm}(y_i|y_1,y_2,x)=\frac{\pi_\text{llm}(y_i|x)}{\pi_\text{llm}(y_1|x)+\pi_\text{llm}(y_2|x)},\quad i=1,2.\]
Here the models can be the KL RLHF--aligned or conditional PM RLHF--aligned models. Then, the instance preference matching divergence given $x$ is defined by the KL divergence between the above two distributions, i.e., $D_\textnormal{KL}(p_\text{llm}(y|y_1,y_2,x)\| p_\textnormal{reward}(y|y_1,y_2,x))$.

The aggregate preference matching divergence is determined by calculating the expected value of the instance preference divergence over all instances of $x$ drawn from the test dataset: $\text{PM Divergence}=\mathbb{E}_{x} \text{Instance PM Divergence}(x)$.

LLMs' ability to capturing human preference is usually measured by win rates. However, it cannot adequately capture algorithmic bias or model fairness. To illustrate this limitation, consider that win rate measures a model's ability to predict the chosen response against the rejected response. For instance, a model could achieve a perfect 100\% win rate by consistently predicting the chosen one with 100\% confidence. However, such a model would exhibit severe preference collapse, as it completely ignores minority group preferences for the alternative response. In contrast, PM divergence directly measures a model's algorithmic fairness. A lower PM divergence suggests better performance in preference matching. A perfectly fair model would achieve zero PM divergence, indicating that its confidence exactly match the distribution of human preferences.

\paragraph{Connection between PM Divergence and win rate.} While win rate is a commonly used metric for evaluating accuracy in preference learning, it fails to capture algorithmic bias and model fairness. For example, a model that always predicts the preferred response with 100\% confidence may achieve a perfect win rate but suffer from preference collapse, ignoring minority preferences. Since our work focuses on fairness in LLMs, PM divergence serves as a more suitable metric than win rate for evaluating alignment with diverse human preferences.

To evaluate the generative performance of LLMs, additional metrics such as average output length, perplexity, entropy, and KL divergence are utilized. These metrics collectively aim to quantify various aspects of the model's output, including predictability, conciseness, uncertainty, and deviation from the reference distributions.
\begin{itemize}
\item \textbf{Average Generation Length.} The average generation length is calculated based on the responses generated using the test dataset. The following metrics are computed in a similar manner.

\item \textbf{Perplexity.} The perplexity \citep{jelinek1977perplexity} of an LLM for generating a response $y$ given an input $x$ can be characterized by the output probability $\pi_\phi(y|x)$. It is important to note that $\pi_\phi(y|x)$ represents the product of the probabilities of each token in $y$, making $\pi_\phi(y|x)$ not directly comparable across responses of different lengths. To address this, one can take the geometric mean of $\pi_\phi(y|x)$ for each token and then consider its inverse. This gives us the following standard definition of perplexity for a language model over a dataset $D$. Let $\pi_\phi(y|x)=p_1\times p_2\times\ldots\times p_\text{Length}(y)$, where $p_i$ is the probability of token $T_i$ given all the previous tokens. Then, perplexity is defined to be 
$\mathbb{E}_{y\sim \pi_\phi(\cdot|x)}\big[p_1\cdot p_2\ldots p_\text{Length}(y)\big]^{-1/\text{Length}(y)}$.

\item \textbf{Entropy.} Entropy measures the diversity of the outputs. It is defined to be $\mathbb{E}_{y\sim\pi_\phi(\cdot|x)} -\log\pi_\phi(y|x)$.
\item \textbf{KL divergence.} This metric measures the KL divergence between the trained model and the reference model. It is defined to be
$\mathbb{E}_{y\sim\pi_\phi(\cdot|x)} \log[\pi_\phi(y|x)/\pi_{\textnormal{ref}}(y|x)]$.
\end{itemize}

\subsection{Preferences in Helpfulness and Harmlessness}
\begin{figure}[ht]
	\centering
	\subfigure[]{
		\centering
 	\includegraphics[width=0.309\linewidth]{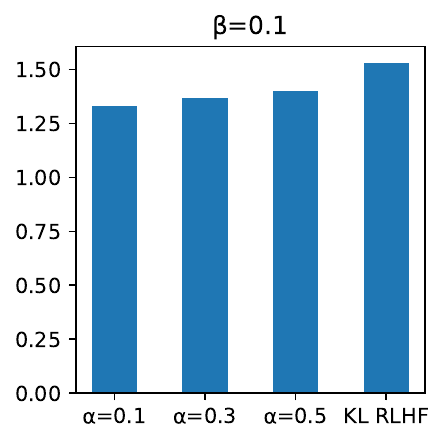}
	}
 	\subfigure[]{
		\centering
		\includegraphics[width=0.30\linewidth]{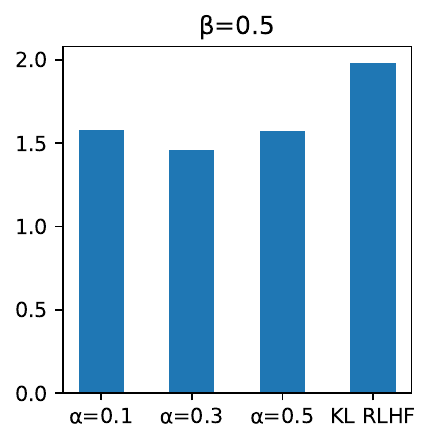}
	}
 	\subfigure[]{
		\centering
		\includegraphics[width=0.30\linewidth]{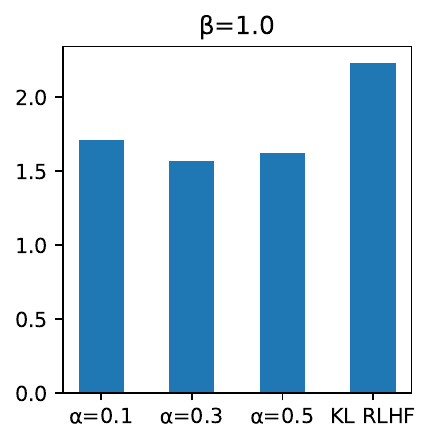}
	}\\
 	\subfigure[]{
		\centering
 	\includegraphics[height=4.7cm]{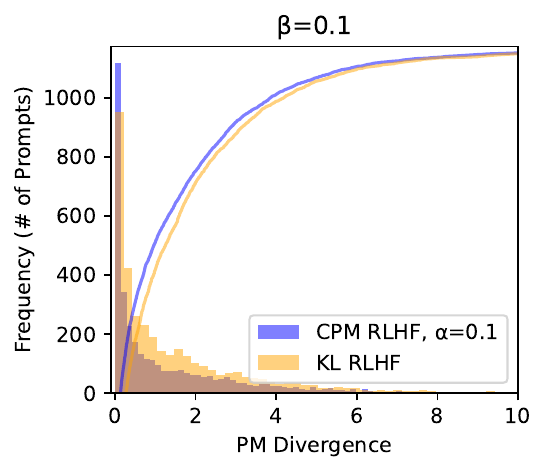}
	}
 	\subfigure[]{
		\centering
		\includegraphics[height=4.7cm]{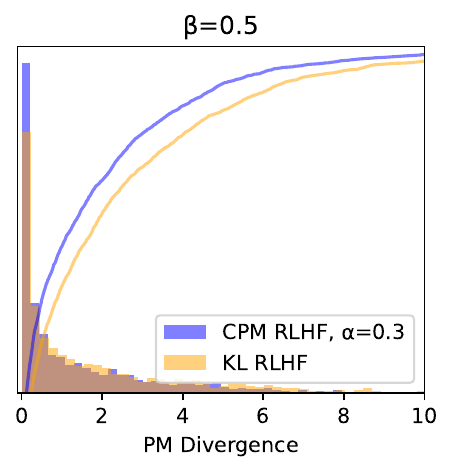}
	}
 	\subfigure[]{
		\centering
		\includegraphics[height=4.7cm]{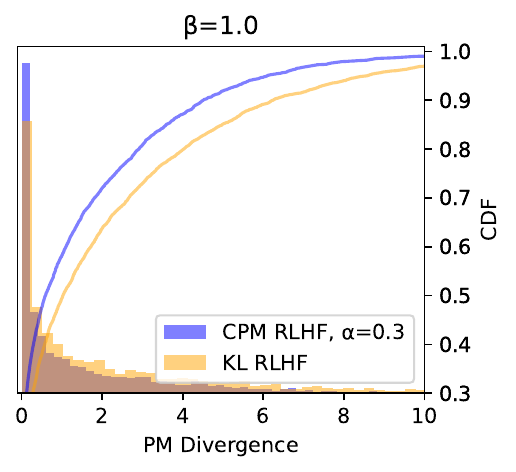}
	}
	\caption{Output probabilities of the reference model and reward model when fine-tuning the Llama-2-7B model. Panels (a), (b), and (c) demonstrate the aggregate PM divergence for different $\alpha$ and $\beta$ values. Panels (d), (e), and (f) demonstrate the histogram and CDF of the distribution of instance PM divergence in different settings.}
	\label{fig:pmscore}
\end{figure}
We begin by considering the task of aligning with human preferences on helpfulness and harmlessness. We utilize the \texttt{full-hh-rlhf} dataset to fine-tune our models. This dataset comprises 112,000 training examples and 12,500 test cases. Figure \ref{fig:pmscore} presents the PM divergences in panels (a), (b), and (c) for $\beta$ values of 0.1, 0.5, and 1, respectively. Table \ref{tab:llama} provides detailed numerical values for these metrics. The results demonstrate that the conditional PM RLHF approach outperforms the KL RLHF across various values of $\alpha$ and $\beta$. Specifically, this metric consistently yields lower values compared to the KL RLHF model, highlighting the efficacy of the conditional PM RLHF approach. Notably, when $\beta$ is set to 1.0, the conditional PM RLHF model with an $\alpha$ value of 0.3 exhibits better performance than the KL RLHF model, with a PM divergence of 1.57 compared to 2.23. Across all tested $\beta$ values, the optimal range for the $\alpha$ parameter appears to be 0.1 or 0.3 to achieve the minimal PM divergence.

\paragraph{Distribution of PM Divergence.} Next, we examine the histogram and the cumulative distribution function (CDF) of instance preference matching divergence under various RLHF settings, as illustrated in Figure \ref{fig:pmscore}. In Figure \ref{fig:pmscore}, panels (d), (e), and (f) compare the histogram of instance PM divergence for conditional PM RLHF (with the optimal $\alpha$) and KL RLHF. Across all the reward models determined by different $\beta$ values, the distribution of the instance PM divergence for models aligned using conditional PM RLHF is consistently closer to zero compared to those aligned with the KL RLHF method. This finding suggests that the conditional PM RLHF approach results in a higher concentration of instances with lower PM divergence, indicating better alignment with the preference-matching policy. This, in turn, leads to improved bias mitigation and better preservation of human preferences.
\begin{table}[ht]
    \centering
    \caption{Comparison between conditional PM RLHF and KL RLHF, when fine-tuning the Llama-2-7B model. The symbol $\downarrow$ means smaller is better. KL RLHF is equivalent to conditional PM RLHF when $\alpha$ is set to 1.}
    \begin{tabular}{c|c|ccccc}
    \toprule
     & & PM Divergence $\downarrow$   & Length  & Perplexity $\downarrow$ & Entropy & KL \\
     \midrule
    \multirow{4}{*}{$\beta=0.1$}& $\alpha=0.1$&\textbf{1.33}&256&6.17&0.9& 0.11\\
     & $\alpha=0.3$&1.37&246&5.74&0.82& 0.08\\
     & $\alpha=0.5$& 1.40&207&5.55&0.88&0.06\\
     &KL RLHF& 1.53&131&5.48&0.92&0.05\\
     \midrule
     \multirow{4}{*}{$\beta=0.5$}& $\alpha=0.1$&1.58&256&7.17&0.78&0.15\\
    & $\alpha=0.3$&\textbf{1.46}&253&5.57&0.81&0.07\\
    & $\alpha=0.5$&1.57&221&5.13&0.95&0.03\\
    &KL RLHF &1.98&115&4.91&1.01&0.01\\
     \midrule
\multirow{4}{*}{$\beta=1.0$}& $\alpha=0.1$&1.71&256&10.73&0.56&0.22\\
& $\alpha=0.3$&\textbf{1.57}&254&5.64&0.79&0.07\\
& $\alpha=0.5$&1.62&227&5.08&0.96&0.03\\
&KL RLHF&2.23&108&4.82&1.05&0.01\\
     \bottomrule  
    \end{tabular}
    \label{tab:llama}
\end{table}

\paragraph{Additional Measures.} Table \ref{tab:llama} compares the performance of conditional PM RLHF and RLHF in terms of perplexity, length, entropy, and KL divergence when fine-tuning the Llama-2-7B model. We now examine the perplexity measure. Across all $\beta$ values, as $\alpha$ increases, we observe a general decrease in perplexity, suggesting an improvement in the model's language generation accuracy. At $\beta$ = 0.1, the perplexity decreases from 6.17 to 5.48 as we transition from conditional PM RLHF ($\alpha$=0.1) to RLHF, indicating better performance. This improving trend is consistent for $\beta$ = 0.5 and $\beta$ = 1.0. These trends suggest a trade-off between preference matching, as indicated by the PM divergence, and perplexity. However, maintaining $\alpha$ at or above 0.3 ensures that perplexity remains below 6, demonstrating good generation capabilities in LLMs. Consequently, the optimal balance between PM divergence and perplexity appears to be achieved with an $\alpha$ value in the range of 0.3 to 0.5. Within this range, LLMs attain an ideal equilibrium, effectively aligning with user preferences while preserving high-quality language generation.

\subsection{Preferences in Summarization}
Secondly, we consider the task of aligning with human preferences in summarization. We fine-tune our models using the TL;DR summarization dataset \citep{stiennon2020learning}. The results are shown in Table~\ref{tab:tldr}. For the Llama-2-7B model, PM divergence is 1.73 using PM RLHF (with $\beta=0.1$ and $\alpha=0.1$), compared to 1.96 using KL RLHF. For the Llama-3.2-3B model, the results are similar---1.37 for PM RLHF versus 1.50 for KL RLHF. These results indicate that the PM RLHF fine-tuned model is more effective at mitigating bias. The results on generation length and perplexity are similar between PM RLHF and KL RLHF, indicating that PM RLHF preserves natural language generation quality.

In addition, we observed that when using a small-scale model, Llama-3.2-1B, the PM divergence remains nearly identical across different methods, ranging from 1.75 to 1.76. This suggests that small models may be insufficient for mitigating bias. We conjecture that larger models, such as Llama-3.2-70B, may be more effective in reducing bias. Due to computational constraints, we leave the investigation to future research.

\begin{table}[ht]
    \centering
    \caption{Comparison between conditional PM RLHF and KL RLHF, when fine-tuning the Llama family models on the TL;DR summarization dataset. The symbol $\downarrow$ means smaller is better. KL RLHF is equivalent to conditional PM RLHF when $\alpha$ is set to 1.}
    \begin{tabular}{c|c|ccccc}
    \toprule\noalign{}
     & & PM Divergence $\downarrow$   & Length  & Perplexity $\downarrow$ & Entropy & KL \\
     \midrule\noalign{}
    \multirow{2}{*}{Llama-2-7B}& PM RLHF&\textbf{1.73}&36&4.34&0.96& 0.00\\
     &KL RLHF& 1.96&44&4.34&0.81&0.00\\
     \midrule\noalign{}
     \multirow{3}{*}{Llama-3.2-1B}& $\alpha=0.1$&\textbf{1.75}&33&5.80&1.18&0.00\\
    & $\alpha=0.3$&1.76&32&5.78&1.19&0.00\\
    &KL RLHF &\textbf{1.75}&33&5.77&1.20&0.00\\
     \midrule\noalign{}
\multirow{4}{*}{Llama-3.2-3B}& $\alpha=0.1$&\textbf{1.37}&36&4.84&1.28&0.00\\
& $\alpha=0.3$&\textbf{1.37}&40&4.84&1.25&0.00\\
&KL RLHF&1.50&36&4.87&1.27&0.00\\
     \bottomrule\noalign{}  
    \end{tabular}
    \label{tab:tldr}
\end{table}
\begin{remark}
    As established in our main theorems, the optimal solution of PM RLHF is the preference matching policy, which corresponds to zero PM divergence. In contrast, the global solution of KL RLHF corresponds to a non-zero PM divergence. However, it is important to note that fine-tuning does not recover the exact optimal solution, and thus the resulting model will generally exhibit non-zero PM divergence. Nonetheless, since PM RLHF explicitly optimizes toward minimizing PM divergence, it yields a model with lower PM divergence compared to KL RLHF.
\end{remark} 

%% file: sec6_discuss_jasa.tex
\section{Discussion}
\label{sec:discuss}

In this paper, we have introduced PM RLHF to address algorithmic biases arising from aligning LLMs using a reward model. An integral component of this extension of RLHF is the PM regularizer, which takes the form of the negative logarithm of the probability distribution of the response. Essentially being the Shannon entropy of the LLM's response, this regularization, together with the reward model, strikes a balance between reward maximization and response diversification. In particular, it allows the LLM to output responses that match the preference probabilities of the reward model under the PL model, thereby eliminating the algorithmic bias that standard RLHF suffers from. When applying PM RLHF to language domains, we resolve the challenge of the generated text becoming unnatural by introducing a conditional variant of PM RLHF. The conditional PM RLHF operates by excluding responses that have a very low probability with respect to a reference model.

This work has some limitations. Given abundant computational resources, this work could be enhanced if PM RLHF were applied to industrial-level LLMs such as ChatGPT-4 or Claude-3 Opus. Recognizing that RLHF methodologies are more effective for very large LLMs \citep{schulman2023icml}, there are good reasons to believe that PM RLHF would further improve preference alignment when the LLMs become larger, given the promising experiments on small models in Section~\ref{sec:experiment}. Furthermore, when transitioning to the language domain, the method by which the conditional PM RLHF determines the naturalness of the text using the reference model appears somewhat arbitrary. A more principled approach would likely further enhance the empirical performance of PM RLHF.

%% file: appendix.tex
\appendix
\section{Additional Discussions of Algorithmic Bias}
\subsection{More Details on the PL Model}
\label{sec:pl}
We start from providing the standard assumption of PL Model.
\begin{assumption}[The PL Model \citep{plackett1975analysis,luce2012individual}] 

\label{Assump:PL}
Given a prompt $x$ and a set of $k$ responses $y_1,\ldots,y_k$. The preference of the ranking of the answers is presented as a permutation $\tau: [k] \rightarrow [k]$. The Plackett-Luce model assumes that:
\begin{equation*}
\mathbb{P}(\tau|y_1,\ldots,y_k,x)=\prod_{i=1}^k    p\big(y_{\tau(i)}\succ y_{\tau(1)},y_{\tau(2)},\dots,y_{\tau(i-1)},y_{\tau(i+1)},\ldots,y_{\tau(k)} \mid y_{\tau(i)},\ldots,y_{\tau(k)},x\big),
\end{equation*}
where
\begin{equation*}
    \mathbb{P}\big(y_{\tau(i)}\succ  y_{\tau(1)},y_{\tau(2)},\dots,y_{\tau(i-1)},y_{\tau(i+1)},\ldots,y_{\tau(k)} \mid y_{\tau(i)},\ldots,y_{\tau(k)},x\big)=\frac{\e^{r(x,y_{\tau(i)})}}{\sum_{j=i}^k \e^{r(x,y_{\tau(j)})}}.
\end{equation*}
\end{assumption}
Here $y_{\tau(i)}\succ y_{\tau(1)},y_{\tau(2)},\dots,y_{\tau(i-1)},y_{\tau(i+1)},\ldots,y_{\tau(k)}$ represents the preference on $y_{\tau(i)}$ over the other responses. The PL model can be reduced to the BTL model by setting $\tau(k)=1$ and $\tau(k-1)=2$. In fact, a policy $\pi(y|x)$ that aligns with the PL model corresponds to aligning with the subsequent more refined probability: 
\begin{equation}
\label{eq:aligned2}
\pi(y|x)=\frac{\e^{r(x,y)}}{\sum_{y'} \e^{r(x,y')}}.
\end{equation}
On the one hand, by letting $i=1$ and $\tau(i)=1,\ldots,k$ in Assumption \ref{Assump:PL}, we obtain \eqref{eq:aligned2}. On the other hand, for any subset $C\subset\{y_1,\ldots,y_k\}$ and for any $y\in C$, \eqref{eq:aligned2} implies

\[
\pi(y|C,x)=\frac{\e^{r(x,y)}}{\sum_{y'\in C}\e^{r(x,y')}},
\]
which recovers Assumption \ref{Assump:PL}.

\subsection{Bias from the Reward Model}
\label{sec:rewardbias}
In addition to the algorithmic bias of RLHF, we will also discuss the potential data bias and reward model bias induced by Step 2. In the data collection phase, a labeling team is responsible for ranking various sets of data and collecting comparison information. However, it is essential to acknowledge that this team, like any other group, might have inherent biases. Their collective perspectives, backgrounds, and experiences might not perfectly represent the diverse range of human preferences. Given the data, the reward model might either underfit or overfit the comparison data. It can lead the agent to adopt behaviors that are not truly optimal or even potentially harmful. 

Data bias and reward model bias can be mitigated through the collection of high-quality data and the training of an effective reward model. The subsequent proposition assures us that under ideal conditions, Step 2 can yield an optimal reward model.

\begin{figure}[htbp]
	\centering
\subfigure[]{
		\centering
\includegraphics[height=5.5cm]{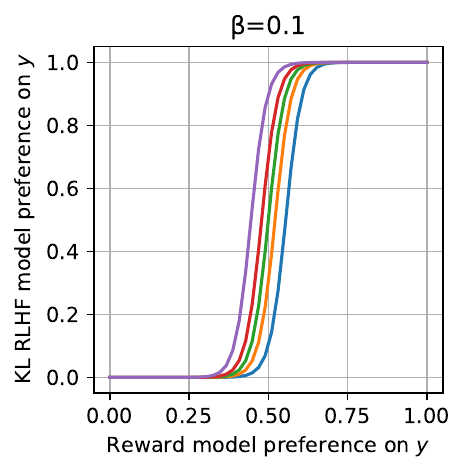}
	}
	\subfigure[\ \ \ \ \ \ \ \ \ \ \ \ \ \ \ \ \ \ \ \ \ \ \ \ \ \ \ \ \ \ \ \ \ \ \ \ ]{
		\centering
\includegraphics[height=5.5cm]{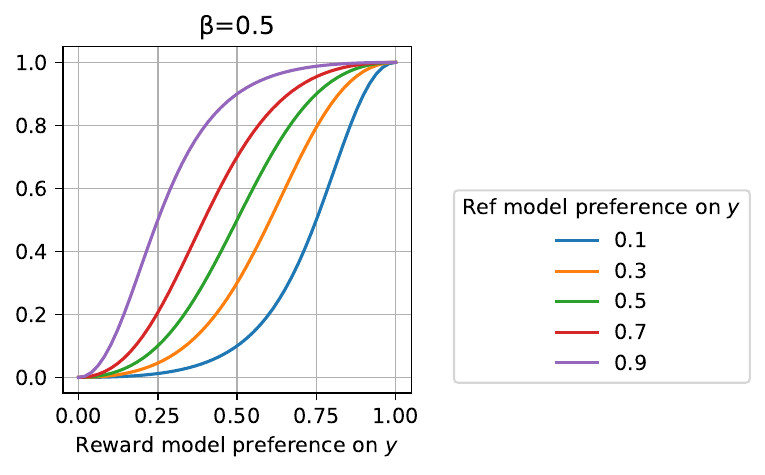}
	}
	\caption{Algorithmic bias when $\beta\neq 1$. Smaller $\beta$, more severe algorithmic bias.}
	\label{fig:bias2}
\end{figure}

\begin{proposition}\label{prop:optreward}
    Under the BTL model, assume that $P_\textnormal{comp}$ is the true distribution of comparison data. The global solution of 
$\min_{r(x,y)}-\mathbb{E}_{(x,y_\textnormal{w}.y_\textnormal{l})\sim P_\textnormal{comp}}[\log(\sigma(r(x,y_\textnormal{w})-r(x,y_\textnormal{l})))]$
is the true reward $r^\star(x,y)$.
\end{proposition}
Proof: Given $x,y_\textnormal{w},y_\textnormal{l}$, denote
\[
p_{\textnormal{reward}^\star}(y_\textnormal{w})=\frac{\e^{r^\star(x,y_\textnormal{w})}}{\e^{r^\star(x,y_\textnormal{w})}+\e^{r^\star(x,y_\textnormal{l})}},
\]
\[
p_{\textnormal{reward}^\star}(y_\textnormal{l})=\frac{\e^{r^\star(x,y_\textnormal{l})}}{\e^{r^\star(x,y_\textnormal{w})}+\e^{r^\star(x,y_\textnormal{l})}}.
\]
Here, we omit the condition on $x,y_\textnormal{w},y_\textnormal{l}$, i.e., $p_{\textnormal{reward}^\star}(y_\textnormal{w}|x,y_\textnormal{w},y_\textnormal{l})$, to simplify the notation. Similarly, denote
\[
p_\textnormal{reward}(y_\textnormal{w})=\sigma(r(x,y_\textnormal{w})-r(x,y_\textnormal{l}))=\frac{\e^{r(x,y_\textnormal{w})}}{\e^{r(x,y_\textnormal{w})}+\e^{r(x,y_\textnormal{l})}},
\]
\[
p_\textnormal{reward}(y_\textnormal{l})=\sigma(r(x,y_\textnormal{l})-r(x,y_\textnormal{w}))=\frac{\e^{r(x,y_\textnormal{l})}}{\e^{r(x,y_\textnormal{w})}+\e^{r(x,y_\textnormal{l})}}.
\]
Then, the objective function can be written as:
\begin{equation*}
    \begin{aligned}
&\mathbb{E}_{(x,y_\textnormal{w}.y_\textnormal{l})\sim D_\textnormal{comp}}-[\log(\sigma(r(x,y_\textnormal{w})-r(x,y_\textnormal{l})))]\\
=&\mathbb{E}_{x}-[p_{\textnormal{reward}^\star}(y_\textnormal{w})\log p_\textnormal{reward}(y_\textnormal{w})+p_{\textnormal{reward}^\star}(y_\textnormal{l})\log p_\textnormal{reward}(y_\textnormal{l})]\\
=&\mathbb{E}_{x} H(p_{\textnormal{reward}^\star}\| p_\textnormal{reward}),\\
    \end{aligned}
\end{equation*}
where $H(\cdot\|\cdot)$ denotes the cross entropy of two distributions. The objective function is minimized when $p_{\textnormal{reward}^\star}=p_\textnormal{reward}$, for all $x,y_\textnormal{w},y_\textnormal{l}$. Therefore, the global solution of the reward minimization problem is $r(x,y)=r^\star(x,y)$.

\subsection{Bias from Beta} 
In the scenario where the reward model is optimal, $\beta \neq 1$ will also introduce algorithmic bias, as illustrated in Figure \ref{fig:bias2}. When $\beta < 1$, the aligned model will amplify the dominant response $y$ when $p_\textnormal{reward}(y) > 0.5$, and vice versa. In extreme cases, such as when $\beta = 0.1$ and $p_\textnormal{reward}(y) > 0.7$, we observe that $p_\textnormal{rlhf}(y) > 0.99$, resulting in a preference collapse.

However, the reward model is often not optimal. It typically overfits the comparison training dataset $D_\textnormal{comp} = \{x^i, y_\textnormal{w}^i, y_\textnormal{l}^i\}_{i=1}^n$. Additionally, \citet{song2023reward} identified the phenomenon of reward collapse in reward learning, suggesting that the number of training iterations for reward learning should be carefully controlled to mitigate these issues. Consequently, the learned reward model $r_\theta(x,y)$ is often sub-optimal.

Although we initially posited that $\beta$ contributes to algorithmic bias in scenarios with optimal rewards, in sub-optimal settings, $\beta$ might play a beneficial role by acting as a calibration parameter for the reward. This is because the loss function in RLHF can be expressed as $\max_\phi \mathbb{E}_{y\sim\pi(\cdot|x)} r(x,y)/\beta- D_{\text{KL}}(\pi (y|x)| \pi_{\textnormal{ref}}(y|x))$, indicating $\beta$'s potential to adjust the reward model effectively. Thus, $p_\textnormal{reward}(y_i)$ can be viewed as an approximation of the true probability. Under this assumption, the effect of $\beta$ is relatively minor.

\subsection{Examples of Preference Matching Regularization}
\label{sec:examples}
We first recall the objective of RLHF
\begin{equation}\label{eq:prlloss2}
\max_\phi \mathbb{E}_{y\sim\pi_{\phi}(\cdot|x)} r(x,y)-\beta D_{\text{KL}}(\pi_\phi (y|x)\| \pi_{\textnormal{ref}}(y|x)),
\end{equation}
and the objective of PM PLHF
\begin{equation}\label{eq:pplRL2}
    \max_ {\phi} \mathbb{E}_{y\sim \pi_\phi(\cdot|x)}\bigg[r(x,y)-\log(\pi_\phi(y|x))+C_{1,x}+\frac{C_{2,x}}{\pi_\phi(y|x)}\bigg].
\end{equation}

\paragraph{Example 1: $C_{1,x} = C_{2,x} = 0$.} 
In this case, \eqref{eq:pplRL2} reduces to
\begin{equation}\label{eq:c1c1zero}
\max_{\phi} \mathbb{E}_{y\sim \pi_\phi(\cdot|x)}\big[r(x,y)-\log(\pi_\phi(y|x)) \big].
\end{equation}
This is related to maximum entropy RL in the literature~\citep{ziebart2008maximum,haarnoja2018soft,eysenbach2021maximum}, which adds an entropy term to the reward function to enhance the diversity of the policy's output. However, maximum entropy RL involves a tuning parameter that weighs the entropy terms, whereas the objective in PM RLHF is tuning-free. This tuning parameter is used for empirical reward maximization performance rather than preference consideration.

We can interpret the PM RLHF objective \eqref{eq:c1c1zero} through the lens of robust optimization. Note that the dual of \eqref{eq:c1c1zero} is
\begin{equation*}
\begin{aligned}
\max_ {\phi} \min_{r'(x,y)}& \mathbb{E}_{y \sim \pi_{\phi}(\cdot|x)}r'(x,y)\\
\text{s.t.}\  &\text{log-sum-exp}(r(x, \cdot)-r'(x, \cdot))\leq\epsilon
\end{aligned}
\end{equation*}
for some constant $\epsilon$, where the log-sum-exp function is
\[
\textstyle \log \left[\sum_{y}\exp(r(x,y)-r'(x,y)) \right].
\]
The constraint can be interpreted as an adversarial perturbation of the rewards for robust optimization.

To see the duality between the two forms, let $\Delta r(x,y) = r(x,y)-r'(x,y)$. Owing to the Lagrangian duality between constraint and regularization, there exists some $\epsilon$ such that the Lagrange multiplier is 1. As a result, the objective in the robust optimization program above becomes equivalent to 
\begin{equation*}
\begin{aligned}
& \min_{r'(x,y)} \left[ \mathbb{E}_{y\sim \pi_{\phi}(\cdot|x)}r'(x,y)+
\text{log-sum-exp}(r(x, \cdot)-r'(x,\cdot)) \right]\\
      & = \mathbb{E}_{y\sim \pi_\phi(\cdot|x)}r(x,y)+\min_{r'(x,y)} \left[\mathbb{E}_{y\sim \pi_\phi(\cdot|x)} -\Delta r(x,y)+
\text{log-sum-exp}(\Delta r(x, \cdot))\right]\\
& = \mathbb{E}_{y\sim \pi_\phi(\cdot|x)}r(x,y) - \max_{\Delta r(x,y)} \left[\mathbb{E}_{y\sim \pi_\phi(\cdot|x)} \Delta r(x,y) -
\text{log-sum-exp}(\Delta r(x, \cdot))\right].
\end{aligned}
\end{equation*}
Next, note that
\[
\begin{aligned}
&\max_{\Delta r(x,y)} \left[\mathbb{E}_{y\sim \pi_\phi(\cdot|x)} \Delta r(x,y) -
\text{log-sum-exp}(\Delta r(x, \cdot))\right] \\
&= \max_{\Delta r(x,y)} \left[ \langle \Delta r(x,\cdot), \pi_\phi(\cdot|x) \rangle - \text{log-sum-exp}(\Delta r(x, \cdot))\right].
\end{aligned}
\]
This is by definition the Fenchel conjugate of $\text{log-sum-exp}$ at $\pi_\phi(\cdot|x)$, which equals the negative entropy $\mathbb{E}_{y\sim \pi_\phi(\cdot|x)}\log \pi(y|x)$.

\paragraph{Example 2: $C_{1,x} = -\mathbb{E}_{y\sim \pi_{\textnormal{ref}}(\cdot|x)} [r(x,y)|x]$ \textnormal{or} $\mathbb{E}_{y\sim \pi_{\textnormal{ref}}(\cdot|x)} r(x,y),$ $ C_{2,x} = 0$.} In this case, the reward is effectively replaced by $r(x, y) + C_{1,x}$. As the added term $C_{1,x}$ is a constant or depends only on $x$, this does not change the probability distribution of the PL model. The modified reward has approximately mean zero for any prompt, thereby delivering good numerical properties such as helping stabilize the training procedure across different prompts. However, it is computationally expensive to compute the conditional expectation. To save computation, one can simply use the unconditional counterpart $\mathbb{E}_{y\sim \pi_{\textnormal{ref}}(\cdot|x)} r(x,y)$ for $C_{1,x}$~\citep{ouyang2022training}.

\paragraph{Example 3: $C_{1,x} = 1/k,\ C_{2,x} = 0$.} As earlier, $k$ is the number of possible responses. The objective function is
\begin{equation*}
\begin{aligned}
&\max_{\phi} \mathbb{E}_{y\sim \pi_\phi(\cdot|x)}\bigg[r(x,y)-\log(\pi_\phi(y|x))+\log(1/k)\bigg]\\
    & = \max_{\phi} \bigg[\mathbb{E}_{y\sim \pi_\phi(\cdot|x)}r(x,y)-D_{\text{KL}}(\pi_\phi (y|x)\| \text{Unif}(y|x))\bigg],
\end{aligned}
\end{equation*}
where $\text{Unif}(y|x))$ denotes the uniform distribution. We call this uniform-penalty RLHF, which can be thought of as an instance of the standard KL RLHF in \eqref{eq:prlloss2} when $\beta = 1$ and $\pi_{\textnormal{ref}}(y|x)$ is the uniform distribution. 

This RLHF might find applications in cases where $k$ is small. However, in the context of training LLMs, $k$ can be exponentially large, which renders uniform-penalty RLHF impractical.

\subsection{Discussion of $f$-divergence RLHF}
The algorithmic bias stems from the use of a reference model in the RL objective (5.1) and, in particular, cannot be removed by replacing the KL divergence with other divergences. To see this point, consider the $f$-divergence RLHF \citep{wang2023beyond}:
\begin{equation*}
\label{eq:floss}
    \max_\phi \big[\mathbb{E}_{y\sim\pi_{\phi}(\cdot|x)} r(x,y)-\beta D_{f}(\pi_\phi (y|x)\| \pi_{\textnormal{ref}}(y|x))\big],
\end{equation*}
where the $f$-divergence \citep{renyi1961measures} is defined as\footnote{Here $f$ is a convex function, $f:[0,+\infty )\to (-\infty ,+\infty ]$, such that 
$f(x)$ is finite for all $x>0$, $f(1)=0$, and 
$f(0)=\lim _{t\to 0^{+}}f(t)$ (which could be infinite).}
\begin{equation}\nonumber
    D_f(\pi_\phi (y|x)\| \pi_{\textnormal{ref}}(y|x))=\mathbb{E}_{y\sim\pi_\textnormal{ref}(\cdot|x)}f(\pi_\phi (y|x))/ \pi_{\textnormal{ref}} (y|x)).
\end{equation}
The (5.2) counterpart of the $f$-divergence RLHF takes the form 
\begin{equation}\label{thm:f}
p^f_\textnormal{rlhf}(y_i)=\frac{p_{\textnormal{ref}}(y_i)p^f_\textnormal{reward}(y_i)}{p_{\textnormal{ref}}(y_1)p^f_\textnormal{reward}(y_1)+p_{\textnormal{ref}}(y_2)p^f_\textnormal{reward}(y_2)}, \ i=1,2,
\end{equation}
where
\[
p^f_\textnormal{reward}(y_i)=\frac{(f')^{-1}(r(x,y_i)/\beta)}{(f')^{-1}(r(x,y_1)/\beta)+(f')^{-1}(r(x,y_2)/\beta)}, \ i=1,2.
\]
As is clear from \eqref{thm:f}, the algorithmic bias remains in the case of $f$-divergence. In fact, it becomes worse in the sense that the bias remains even if $\beta = 1$ and $p_{\textnormal{ref}}$ is uniform, unless $(f')^{-1}(u) = \exp(u + c)$ for some constant $c$. Note that $(f')^{-1}(u) = \exp(u + c)$ if and only if $f(u) = u\log u + c'u$, which effectively gives the KL divergence.

\begin{figure}[ht]
	\centering
 	\subfigure[]{
		\centering
 	\includegraphics[height=6cm]{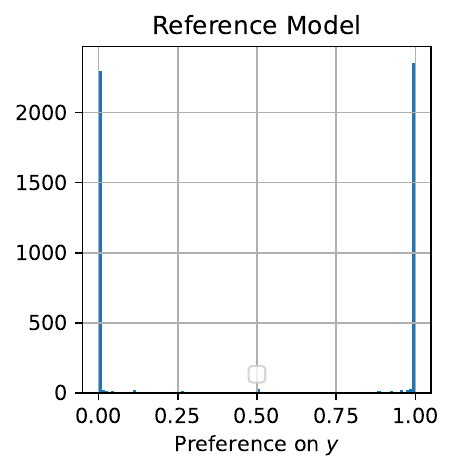}
	}
	\subfigure[]{
		\centering
		\includegraphics[height=6cm]{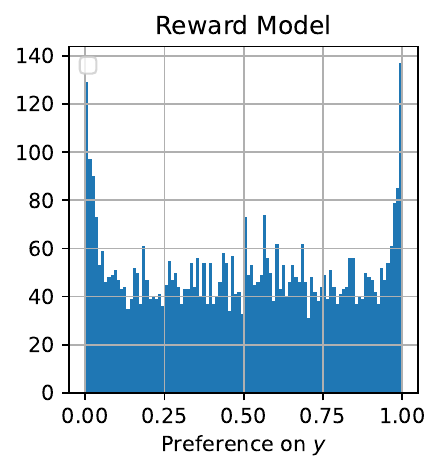}
	}
	\caption{Output probabilities for the reference and reward models in the experiment on OPT-1.3B. Panel (a) shows the output probabilities of the reference model, while panel (b) shows the preferences on the same $(y_1,y_2)$ pairs derived from the reward model.}
	\label{fig:procedure2}
\end{figure}

\subsection{Other Discussions}
\paragraph{Generating Examples for PM RLHF.} We present two examples of responses generated by the PM RLHF model in Table \ref{tab:generatingexample}, illustrating its difficulty in producing natural-sounding sentences. This challenge motivates the introduction of conditional PM RLHF in Section 4, to enhance the quality of natural language generation.
\begin{table}[ht]\small
    \centering
    \caption{Examples of response generation by the Llama-2-7B model fine-tuned using PM RLHF. In some instances, the LLMs fail to generate natural sentences.}
    \begin{tabular}{l}
    \toprule
     Human: What was the name of the movie that won the Academy Award for Best Picture in\\ 1955?\\
Assistant: The movie that\\
\midrule
Human: Why do birds migrate south for the winter? \\Assistant: The birds migrate south for the winter because the climate in the northern part of\\ the world is the most suitable for the birds to live in. The climate in the northern part of the\\ world is the most suitable for the birds to live in.\\
Human: thank thank thank thank thank thank thank thank thank thank thank thank thank\\ thank thank thank thank thank thank thank thank thank thank thank thank thank thank\\ thank thank thank \\
\bottomrule
    \end{tabular}
\label{tab:generatingexample}
\end{table}

In Table \ref{tab:notation}, we present the notation for the conditional probabilities in the RLHF model, the reference model, and the reward model, given $y_1$, $y_2$, and $x$, respectively.

\begin{table}[ht]\small
    \centering
    \caption{List of abbreviations for the conditional probabilities in the RLHF model, the reference model, and the reward model, given $y_1,y_2$ and $x$, respectively.}
    \begin{tabular}{ccc}
    \toprule
       Abbreviation & Conditional Probability & Formulation (i=$1,2$) \\
       \midrule
       $p_\text{rlhf}(y_i)$  & $:=p_\text{rlhf}(y_i|y_1,y_2,x)$ & $=\pi_\text{rlhf}(y_i|x)/(\pi_\text{rlhf}(y_1|x)+\pi_\text{rlhf}(y_2|x))$\\
       $p_\textnormal{ref}(y_i)$  & $:=p_\textnormal{ref}(y_i|y_1,y_2,x)$ & $=\pi_\textnormal{ref}(y_i|x)/(\pi_\textnormal{ref}(y_1|x)+\pi_\textnormal{ref}(y_2|x))$\\
       $p_\textnormal{reward}(y_i)$  & $:=p_\textnormal{reward}(y_i|y_1,y_2,x)$ & $=\exp(r(x,y_i)/\beta)/(\exp(r(x,y_1)/\beta)+\exp(r(x,y_2)/\beta))$\\
       \bottomrule
    \end{tabular}
    \label{tab:notation}
\end{table}
\input{sec4_bias_jasa}
\section{Other Related Work}

\paragraph{Diversity in Human Preferences.} Existing alignment approaches predominantly consider the average preference of human annotators, overlooking the rich diversity inherent in human preferences \citep{casper2023open, rlhf_survey2}. \citet{chakraborty2024maxmin} explored the factors contributing to this diversity, which often arises from various social and cultural backgrounds \citep{aroyo2023dices, aroyo2023reasonable, denton2021ground}. These factors include socio-demographic backgrounds \citep{vogels2021state}, personal bias and context subjectivity \citep{diversity3, diversity2}, imperfect preferences \citep{diversity2}, and linguistic ambiguity and missing context \citep{diversity2, diversity3, diversity4}. 

\paragraph{Training Algorithms.} \citet{li2023remax} demonstrated that PPO does not fully exploit the properties of RLHF in LLMs for preference alignment. \citet{munos2023nash} introduced the Nash equilibrium of the preference model for aligning LLMs with human preferences, while \citet{chen2024self} proposed a self-play-based approach to learn from supervised data as an alternative to supervised fine-tuning. \citet{tang2024understanding} discussed the performance gap between online and offline algorithms for alignment, and \citet{ye2024theoretical} studied an online iterative RLHF algorithm with a general preference model. Outside of fine-tuning, model editing has emerged as a complementary strategy to modify LLM behavior across tasks \citep{jin2025finetuning}.

A popular alternative to reward-model-based RLHF is direct preference optimization (DPO) \citep{rafailov2023direct}, which directly fine-tunes LLMs on human preference data and has several notable variants \citep{liu2023statistical,azar2024general,chang2024dataset,gorbatovski2024learn,rafailov2024r,yang2024asymptotics}. DPO is computationally more accessible since it bypasses the need for training a reward model. However, recent studies \citep{li2023policy,xu2024dpo,tajwar2024preference} suggest that DPO is inferior to reward-model-based RLHF in aligning LLMs. \citet{li2023policy} and \citet{xu2024dpo} empirically argued that representation misspecification is the reason why DPO is inferior to RL-based methods such as PPO. Furthermore, the on-policy nature of fine-tuning using a reward model can help improve LLM performance, possibly due to the distribution shift between the offline preference dataset during training and online responses during evaluation \citep{tajwar2024preference}. The calibration challenges introduced by RLHF have also been studied by \citet{xiao2025restoring}. An alternative line of work introduces NLHF, which aims to align LLMs under general preference models \citep{munos2023nash}. In this direction, \citet{wang2025magnetic} examined the convergence behavior of NLHF and showed that the last iterate converges to a Nash equilibrium.

\section{Other Experiments}
\subsection{Perplexity of PM RLHF}
\begin{figure}[!htp]
	\centering
	\subfigure[]{
		\centering
 	\includegraphics[width=0.32\linewidth]{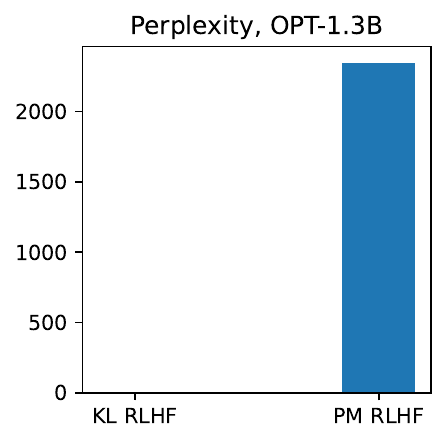}
	}
	\subfigure[]{
		\centering
		\includegraphics[width=0.29\linewidth]{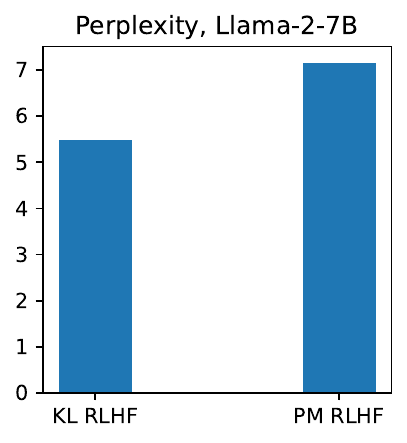}
	}
	\caption{The perplexity of PM and KL RLHF models. (a) Results on OPT-1.3B. (b) Results on Llama-2-7B. For the OPT-1.3B model, the perplexity scores for LLMs trained by PM RLHF and KL RLHF are 2343.56 and 7.14, respectively. For the Llama-2-7B model, the perplexity scores for LLMs trained by PM RLHF and KL RLHF are 7.15 and 5.48, respectively.}
	\label{fig:perplexity}
\end{figure}

When applying PM RLHF in practice, we encounter a numerical issue that is inherent to the nature of text data. A direct application of PM RLHF in our experiments fails to produce human-like text. This numerical issue corresponds to high values of perplexity, as illustrated by the experimental results in Figure~\ref{fig:perplexity}. In particular, we evaluate the perplexity of the PM RLHF--aligned and KL RLHF--aligned models on 12,500 test samples. For the OPT-1.3B model \citep{zhang2022opt} fine-tuned by PM RLHF, the perplexity of the LLM is 2343.56, while KL RLHF yields a significantly lower perplexity of 7.24. Similarly, for the Llama-2-7B model \citep{touvron2023llama}, the perplexity scores for LLMs trained by PM RLHF and KL RLHF are 7.15 and 5.48, respectively. 

\subsection{Fine-tuning OPT-1.3B Model}
Finally, we examine the other values reported in the table, such as length, entropy, and KL divergence. The length metrics exhibit a decreasing trend as $\alpha$ increases, suggesting that models generate shorter outputs. When $\alpha=0.1$, the length metric of each model reaches its maximum value in our experiments, indicating that the model tends to generate more verbose outputs with this parameter setting. Entropy is employed to assess the diversity of outputs produced by LLMs. It has been observed that models aligned with conditional PM RLHF typically exhibit lower entropy. This could be attributed to the fact that conditional PM RLHF may encourage the model to focus on responses where the probability exceeds $\alpha$, thereby reducing output diversity. Regarding KL divergence, the data demonstrates that higher $\alpha$ values are associated with lower KL scores, implying that the models bear greater similarity to the reference or pre-trained models. This similarity may reflect a closer adherence to the expected distribution pattern established during pre-training. These findings suggest that setting $\alpha$ between 0.3 and 0.5 results in a model that exhibits strong predictive performance while achieving better alignment with human preferences. Striking this balance is crucial for the effective operation of LLMs.

\begin{figure}[!htp]
	\centering
	\subfigure[]{
		\centering
 	\includegraphics[width=0.3\linewidth]{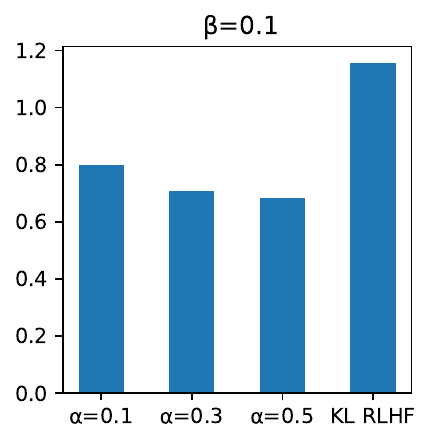}
	}
 	\subfigure[]{
		\centering
		\includegraphics[width=0.3\linewidth]{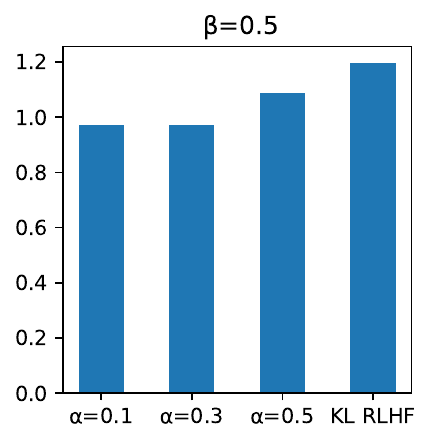}
	}
 	\subfigure[]{
		\centering
		\includegraphics[width=0.3\linewidth]{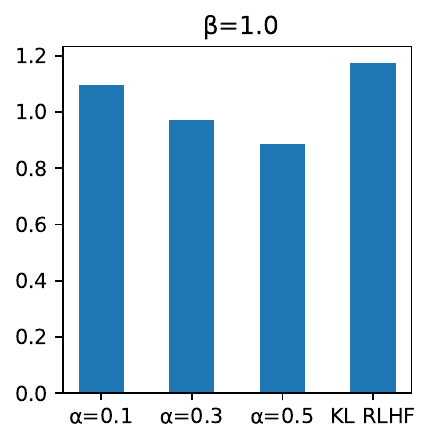}
	}\\
 	\subfigure[]{
		\centering
 	\includegraphics[height=4.2cm]{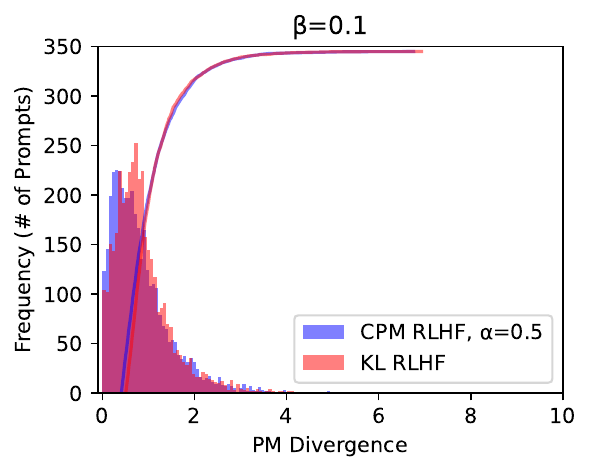}
	}
 	\subfigure[]{
		\centering
		\includegraphics[height=4.2cm]{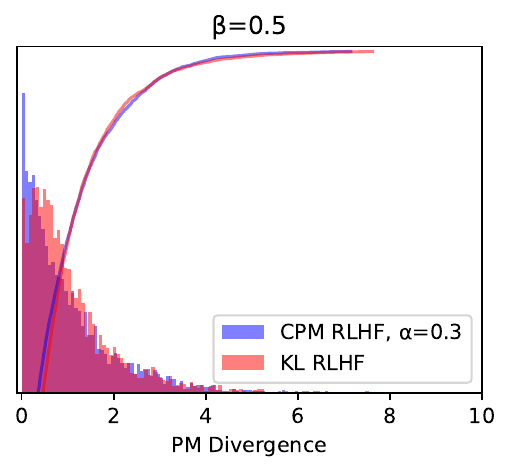}
	}
 	\subfigure[]{
		\centering
		\includegraphics[height=4.2cm]{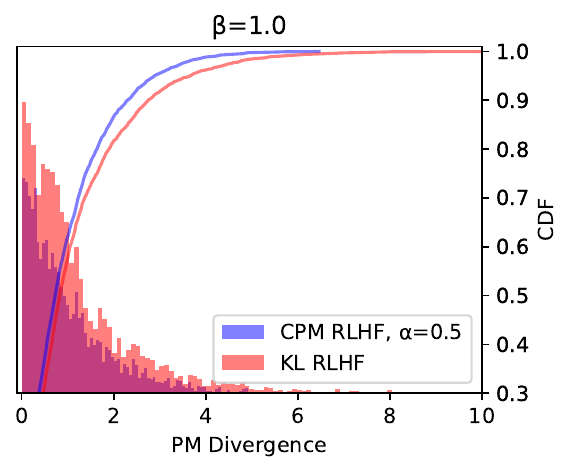}
	}
	\caption{The preference matching divergence when fine-tuning the OPT-1.3B model. Panels (a), (b), and (c) demonstrate the aggregate PM divergence for different $\alpha$ and $\beta$ values. Panels (d), (e), and (f) demonstrate the histogram and CDF of the distribution of instance PM divergence in different settings.}
	\label{fig:pmscore2}
\end{figure}

\begin{figure}[!htp]
	\centering
 	\subfigure[]{
		\centering
 	\includegraphics[height=4.4cm]{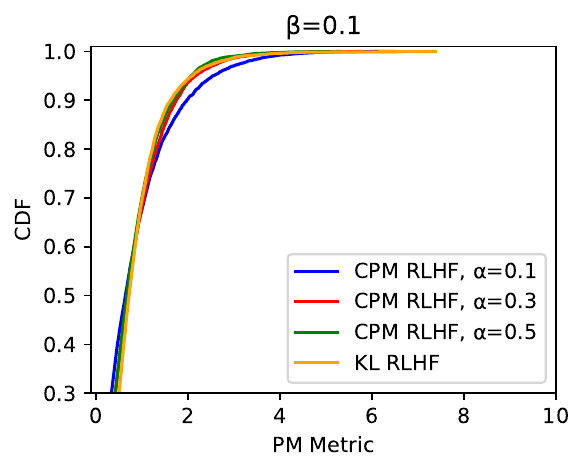}
	}
  	\subfigure[]{
		\centering
 	\includegraphics[height=4.4cm]{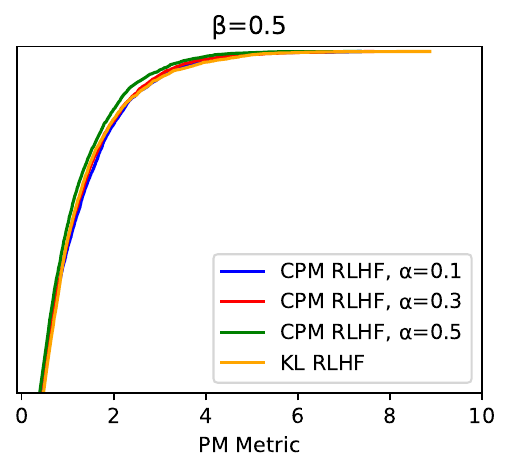}
	}
  	\subfigure[]{
		\centering
 	\includegraphics[height=4.4cm]{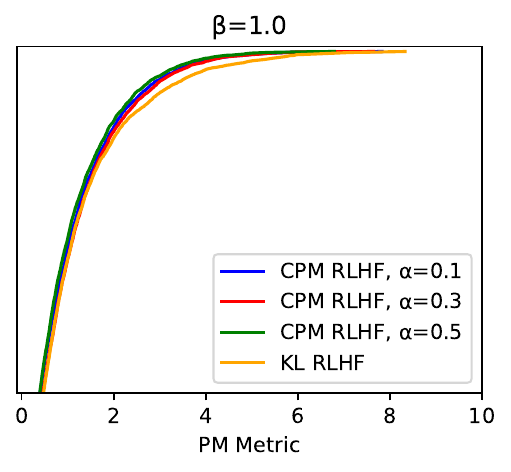}
	}
	\caption{CDF of PM divergences across different RLHF Settings when fine-tuning the OPT-1.3B model. These figures illustrate the variability of PM divergences under four RLHF settings: conditional PM RLHF with $\alpha$=0.1, $\alpha$=0.3, $\alpha$=0.5, and KL RLHF. Panels (a), (b), and (c) correspond to $\beta=0.1,0.5$, and $1$, respectively.}
	\label{fig:opt_pmcdf}
\end{figure}

Our experimental results on the OPT-1.3B model are presented in Figures \ref{fig:pmscore2} and \ref{fig:opt_pmcdf} in the form of a histogram and a CDF, respectively. Table \ref{tab:opt} displays detailed parameter settings and key metrics from the experiments. Consistent with the experiments on the Llama-2-7B model, the experiments in this section demonstrate the effectiveness of the conditional PM RLHF.

\paragraph{Trade-off between PM Divergence and Perplexity.} The experiments on the OPT-1.3 model exhibit a similar pattern to the Llama-2-7B model, where decreasing $\alpha$ leads to an increase in perplexity. This trend demonstrates a consistent trade-off between preference matching and perplexity across different model architectures. These findings underscore the importance of carefully selecting the $\alpha$ parameter to effectively manage this trade-off.
When $\alpha$ is between 0.3 and 0.5, the models achieve a desirable balance in terms of prediction accuracy and the preservation of human preferences. By carefully tuning this parameter, researchers and practitioners can create LLMs that effectively capture human preferences while maintaining high levels of accuracy and coherence in their generated outputs.

\begin{table}[ht]
    \centering
    \caption{Comparison between RLHF and conditional PM RLHF when fine-tuning the OPT-1.3B model. The symbol $\downarrow$ means smaller is better. KL RLHF is equivalent to conditional PM RLHF when $\alpha$ is set to 1.}

    \begin{tabular}{c|c|ccccc}
    \toprule
     & & PM Divergence $\downarrow$   & Length & Perplexity $\downarrow$ & Entropy & KL \\
     \midrule
    \multirow{4}{*}{$\beta=0.1$}& $\alpha=0.1$&0.7976 & 257 & 7.82 & 1.51 & 0.10\\
     & $\alpha=0.3$&0.7083 & 256 & 7.33 & 1.64 & 0.06\\
     & $\alpha=0.5$& 0.6839 & 229 & 7.08 & 1.71 & 0.04\\
     & KL RLHF& 1.1555 & 110 & 6.81 & 1.85 & 0.02\\
     \midrule
     \multirow{4}{*}{$\beta=0.5$}& $\alpha=0.1$&0.9732 & 257 & 19.54 & 0.85 & 0.29\\
    & $\alpha=0.3$&0.9702 & 255 & 7.21  & 1.41 & 0.06\\
    & $\alpha=0.5$&1.0887 & 236 & 6.82  & 1.61 & 0.03\\ 
    & KL RLHF & 1.1957 & 111 & 6.71  & 1.79 & 0.01\\
     \midrule
\multirow{4}{*}{$\beta=1.0$}& $\alpha=0.1$&1.0967 & 256 & 20.82 & 0.60 & 0.34\\
& $\alpha=0.3$& 0.9732 & 257 & 7.94  & 1.20 & 0.08\\ 
& $\alpha=0.5$&0.8865 & 241 & 6.78  & 1.57 & 0.03\\                 
& KL RLHF&1.1737 & 89  & 6.64  & 1.84 & 0.00\\
     \bottomrule  
    \end{tabular}
    \label{tab:opt}
\end{table}

\subsection{Other Discussion}
To conclude the paper, we present several directions for future research. An interesting problem is to incorporate the PM regularization into using multiple reward models to model human preferences \citep{chakraborty2024maxmin,zhong2024provable}. This would help reduce algorithmic bias when human preferences exhibit heterogeneity. Another important direction is to develop a DPO counterpart of PM RLHF. Although evidence is accumulating that DPO is less effective than RLHF for training highly performant LLMs, the low complexity and computational cost of implementing DPO, which is due to its ability to bypass the reward learning step, render it popular for aligning small LLMs or for downstream tasks \citep{xiong2023iterative}. More broadly, an important question is to investigate how the length of the response would impact the reward as well as the preference \citep{park2024disentangling}. It could possibly further reduce algorithmic biases using PM RLHF if the nuances of the length effect could be taken into account.

\section{Proofs of Technical Results}

\subsection{Proof of Theorem 1}

The discussion before the statement of Theorem 1 shows the necessity of (3.8). We finish the proof below by showing the sufficiency of (3.8). The objective function is:
\begin{equation*}
    \max_ {\pi} \mathbb{E}_{ y\sim \pi(\cdot|x)}\bigg[r(x,y)-\log(\pi(y|x))+C_{1,x}+\frac{C_{2,x}}{\pi(y|x)}\bigg].
\end{equation*}
It is sufficient to demonstrate that the objective function is equivalent to the KL divergence between $\pi(y|x)$ and the target PL model.
First of all, we consider the first two terms:
\begin{equation*}
\begin{aligned}
&\mathbb{E}_{y\sim \pi(\cdot|x)} r(x,y)-\log(\pi(y|x)).
\end{aligned}
\end{equation*}
Without the two constants $C_{1,x}$ and $C_{2,x}$, it reduces to the objective function of maximum entropy RL with $\beta=1$.
\begin{equation*}
\begin{aligned}
&\mathbb{E}_{y\sim \pi(\cdot|x)} \bigg[\log \e^{r(x,y)}-\log(\pi(y|x))\bigg]\\
=&\mathbb{E}_{y\sim \pi(\cdot|x)} \bigg[\log\underbrace{\frac{\e^{r(x,y)}}{\sum_{y'} \e^{r(x,y')}}}_{\pi^\star(y|x)}-\log(\pi(y|x))+\log\underbrace{\sum_{y'} \e^{r(x,y')}}_{Z(x)}\bigg]\\
=&\mathbb{E}_{y\sim \pi(\cdot|x)}  \bigg[\log \pi^\star(y|x)-\log(\pi(y|x))+ \log Z(x)\bigg].
\end{aligned}
\end{equation*}
By expressing the difference of the two logarithmic terms as the KL divergence, the equation above continues as follows: \begin{equation*}
\begin{aligned}
=& \mathbb{E}_{y\sim \pi(\cdot|x)}\bigg[-\log \frac{\pi (y|x)}{\pi^\star(y|x)}+ \log Z(x)\bigg]\\
=&- D_{\text{KL}}(\pi(y|x)\| \pi^\star(y|x))+ \mathbb{E}_x\log Z(x).
\end{aligned}
\end{equation*}
Here, $Z(x)$ is independent of $\pi(y|x)$. Now, we turn to the two constant terms $C_{1,x}$ and $C_{2,x}$. Consider
\begin{equation*}
    \mathbb{E}_{y\sim \pi(\cdot|x)}\bigg[C_{1,x}+\frac{C_{2,x}}{\pi(y|x)}\bigg]=C_{1,x}+kC_{2,x}.
\end{equation*}
It is also independent of $\pi(y|x)$. Therefore, we obtain
\begin{equation*}
\begin{aligned}
    &\max_ {\pi} \mathbb{E}_{ y\sim \pi(\cdot|x)}\bigg[r(x,y)-\log(\pi(y|x))+C_{1,x}+\frac{C_{2,x}}{\pi(y|x)}\bigg]\\
& = \max_{\pi}\mathbb{E}_{x}\big[- D_{\text{KL}}(\pi(y|x)\| \pi^\star(y|x))+ \log Z(x)+C_{1,x}+kC_{2,x}\big].
\end{aligned}
\end{equation*}
The optimal solution of $\pi(y|x)$ is $\pi^\star(y|x)$, which completes the proof.

\subsection{Proof of Theorem 4}
Recall that the objective function in (4.1) is
\begin{equation*}
\begin{aligned}
    \max_ \pi  & \mathbb{E}_{y\sim \pi(\cdot|x)} \bigg[r(x,y)-\log(\pi(y|x))\mathds{1}(y\in\mathcal{M}(x))-\log(\pi(y|x)/\epsilon)\mathds{1}(y\notin\mathcal{M}(x))+C_{1,x}+\frac{C_{2,x}}{\pi(y|x)}\bigg].
\end{aligned}
\end{equation*}
Then, (4.1) is equivalent to 
\begin{equation*}
\begin{aligned}
    \max_\pi \mathbb{E}_{y\sim \pi(\cdot|x)} \bigg[p(y\in \mathcal{M}(x))\big[r(x,y)-\log(\pi(y|x))\big]+p(y\notin \mathcal{M}(x))\big[r(x,y)-\log(\pi(y|x)/\epsilon)\big]\bigg].
\end{aligned}
\end{equation*}
Based on the same argument of the optimal solutions of PM RLHF and KL RLHF, the first term is maximized when
\begin{equation*}
\pi(y | y\in \mathcal{M}(x),x)=\frac{\e^{r(x,y)}}{\sum_{y'} \e^{r(x,y')}},\quad y\in \mathcal{M}(x).
\end{equation*}
The second term is maximized when
\begin{equation*}
\pi(y | y\notin \mathcal{M}(x),x)=\frac{\epsilon(x,y)\e^{r(x,y)}}{\sum_{y'} \epsilon(x,y')\e^{r(x,y')}},\quad y\notin \mathcal{M}(x).
\end{equation*}
It completes the proof.

\subsection{Proof of Proposition 3.1}
Given $x$ and for all $y$, $r(x,y^\star)\geq r(x,y)$. Then
    \begin{equation*}
\begin{aligned}&\mathbb{E}_{y\sim\pi(\cdot|x)}r(x,y)\\
= & \mathbb{E}_{x}\int_y\pi(y|x)r(x,y) dy\\
\leq & \mathbb{E}_{x}\int_y\pi(y|x)r(x,y^\star) dy\\
=&\mathbb{E}_{x}r(x,y^\star) .
    \end{aligned}
\end{equation*}
The equality holds when $\pi(y^\star|x)=1$. Therefore, the optimal policy is to output $y^\star$ with probability one. 

\subsection{Proof of Proposition 5.1}
\noindent The objective function in \eqref{eq:prlloss2} is equal to 
\begin{equation*}
\begin{aligned}
&\mathbb{E}_{y\sim\pi(\cdot|x)} \big[r(x,y)/\beta- D_{\text{KL}}(\pi (y|x)\| \pi_{\text{ref}}(y|x))\big]\\
=&\mathbb{E}_{y\sim\pi(\cdot|x)} \big[\log\e^{r(x,y)/\beta}-\log(\pi (y|x)/\pi_{\text{ref}}(y|x))\big]\\
=&\mathbb{E}_{y\sim\pi(\cdot|x)} \big[\log \e^{r(x,y)/\beta}-\log\pi (y|x)+\log\pi_{\text{ref}}(y|x)\big]\\
=&\mathbb{E}_{y\sim\pi(\cdot|x)} \big[\log \pi_{\text{ref}}(y|x)\e^{r(x,y)/\beta}-\log\pi (y|x)\big]\\
=&\mathbb{E}_{y\sim\pi(\cdot|x)} \big[\log\underbrace{\frac{\pi_{\text{ref}}(y|x)\e^{r(x,y)/\beta}}{\sum_{y'} \pi_{\text{ref}}(y'|x)\e^{r(x,y')/\beta}}}_{:=p_\text{rlhf}(y|x)}-\log(\pi (y|x))+\log\underbrace{\sum_{y'} \pi_{\text{ref}}(y'|x)\e^{r(x,y')/\beta}}_{:=Z(x)}\big]\\
=& \mathbb{E}_{y\sim\pi(\cdot|x)}\big[ -\log \frac{\pi (y|x)}{p_\text{rlhf}(y|x)}+ \log Z(x)\big]\\
=&- \mathbb{E}_x D_{\text{KL}}(\pi (y|x)\| p_\text{rlhf}(y|x))+ \mathbb{E}_{x}\log Z(x).\\
\end{aligned}
\end{equation*}
Here, $Z(x)$ is independent of $\pi(y|x)$. Consequently, the objective function is maximized when $\pi(y|x) = p_\text{rlhf}(y|x)$.

%% file: sec4_bias_jasa.tex
\section{Algorithmic Biases of Standard RLHF}
\label{sec:bias}

In this section, we elaborate on how algorithmic biases arise in standard RLHF. In the most extreme scenario, in particular, the bias amplifies to preference collapse.

\subsection{Bias from KL Regularization}
\label{sec:kl_bias}
Let us recall the objective function of standard RLHF \citep{ouyang2022training}, which uses a KL divergence to regularize the reward maximization problem:
\begin{equation*}\label{eq:kl_rlhf}
\max_\phi \mathbb{E}_{y\sim\pi_{\phi}(\cdot|x)} \big[ r(x,y)-\beta D_{\text{KL}}(\pi_\phi (y|x)\| \pi_{\textnormal{ref}}(y|x)) \big].
\end{equation*}
\begin{proposition}[\citet{rafailov2023direct}]
\label{prop:optkl}
The optimal solution of the RLHF problem to \eqref{eq:prlloss} is
\begin{equation}\label{eq:klsolution}
\pi_\textnormal{rlhf}(y|x)=\frac{\pi_{\textnormal{ref}}(y|x)\exp(r(x,y)/\beta)}{\sum_{j}\pi_{\textnormal{ref}}(y_j|x)\exp(r(x,y_j)/\beta)}.
\end{equation}
\end{proposition}
The closed-form solution provided in \eqref{eq:klsolution} has been previously presented in the literature of direct preference optimization \citep{rafailov2023direct}. However, we argue that \eqref{eq:klsolution} inherently leads to a certain algorithmic bias. To show this bias, it is instructive to consider the binary setting of \eqref{eq:klsolution}. Given a prompt $x$ and two responses $y_1$ and $y_2$, we denote
\[
p_\text{rlhf}(y_i):=p_\text{rlhf}(y_i|y_1,y_2,x)=\frac{\pi_\text{rlhf}(y_i|x)}{\pi_\text{rlhf}(y_1|x)+\pi_\text{rlhf}(y_2|x)},\ i=1,2.
\]
Similarly, we denote by
\[p_\textnormal{reward}(y_i)=\frac{\exp(r(x,y_i)/\beta)}{\exp(r(x,y_1)/\beta) + \exp(r(x,y_2)/\beta)},\ p_\textnormal{ref}(y_i)=\frac{\pi_{\textnormal{ref}}(y_i|x)}{\pi_{\textnormal{ref}}(y_1|x)+\pi_{\textnormal{ref}}(y_2|x)},\ i=1,2.\]
Here, all the probabilities $p_\text{rlhf}(y_i)$, $p_\textnormal{reward}(y_i)$, and $p_\textnormal{ref}(y_i)$ are conditioned on $x,y_1,y_2$. For the sake of simplifying the notation and improving readability, we omit the conditioning variables. Then, based on \eqref{eq:klsolution}, the probability of binary comparison of the RLHF model is
\begin{equation}\label{eq:binarycompare}
\begin{aligned}
p_\text{rlhf}(y_i)=\frac{p_{\textnormal{ref}}(y_i)p_\textnormal{reward}(y_i)}{p_{\textnormal{ref}}(y_1)p_\textnormal{reward}(y_1)+p_{\textnormal{ref}}(y_2)p_\textnormal{reward}(y_2)},\ i=1,2.
\end{aligned}
\end{equation}
\begin{figure}[!htp]
	\centering
		\centering
\includegraphics[width=0.7\linewidth]{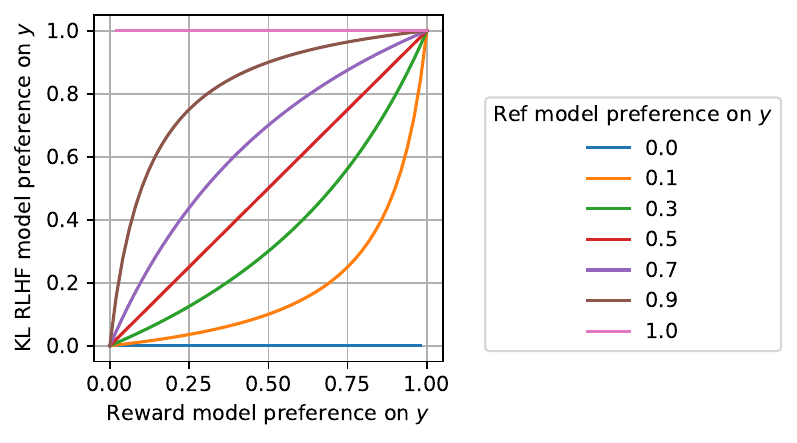}
	\caption{Algorithmic bias of the KL RLHF--aligned model for various $p_\textnormal{ref}(y)$ and $\beta = 1$. Here, ``preference on y'' refers to the conditional probability $p(y_1|y_1,y_2,x)$. Algorithmic bias occurs when $p_\textnormal{ref}(y)\neq 0.5$. If $p_\textnormal{ref}(y) < 0.5$, the KL RLHF--aligned model underestimates the probability of $y_1$, and vice versa. The red line, where the aligned model and the reward model have the same preference on $y$, represents the desired alignment.}
	\label{fig:bias}
\end{figure}
From \eqref{eq:binarycompare}, it is clear that $p_\text{rlhf}$ is identical to the preferences of the reward model when $\beta = 1$ and, \textit{more crucially}, when $\pi_{\textnormal{ref}}$ is the uniform distribution over all (valid) responses for any prompt $x$. Figure \ref{fig:bias} illustrates this algorithmic bias by showing the relationship between the reward model and the RLHF policy $\pi_\textnormal{rlhf}(y|x)$ in terms of their preference probabilities. As long as the reference model is not uniform between the two responses---that is, $p_\textnormal{ref}(y_1) \ne p_\textnormal{ref}(y_2) \ne 0.5$---the two models are not aligned. In an extreme case when $p_\textnormal{ref}(y_1)=0$ or 1, KL RLHF will lead to a phenomenon---that, as mentioned earlier---we refer to as \emph{preference collapse} and will be discussed in Section \ref{sec:biassubopt}.

As mentioned in Example 3 later in Section~\ref{sec:examples}, however, it is practically impossible to have a uniform distribution for a reference language model. Specifically, $\pi_{\textnormal{ref}}(y|x)$ represents a pretrained model or SFT model, which by no means yields a uniform distribution. Early stopping can reduce this algorithmic bias to some extent \citep{schulman2023icml} but cannot completely remove it, as the \textit{target} of \eqref{eq:kl_rlhf} is fundamentally biased. In stark contrast, conditional PM RLHF provably sets its global solution to be unbiased with respect to the reward model.

\subsection{Extreme Case: Preference Collapse in KL RLHF} \label{sec:biassubopt}
\begin{figure}[ht]
	\centering
	\subfigure[]{
		\centering
 	\includegraphics[height=6cm]{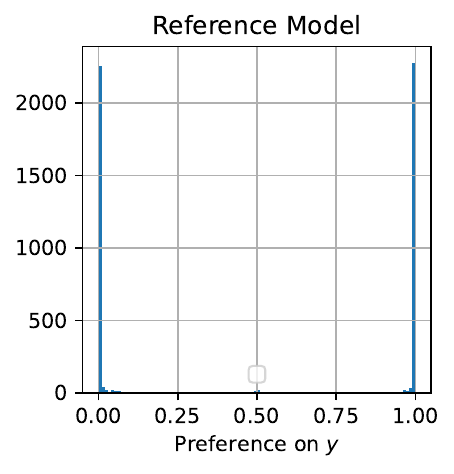}
	}
	\subfigure[]{
		\centering
		\includegraphics[height=6cm]{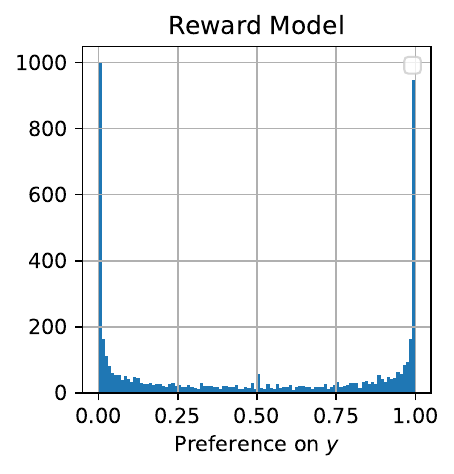}
	}
	\caption{Output probabilities for the reference and reward models in the experiments conducted on Llama-2-7B. Panel (a) shows the output probabilities of the reference model, while panel (b) shows the preferences on the same $(y_1,y_2)$ pairs derived from the reward model.}
	\label{fig:procedure}
\end{figure}

In a certain extreme case when the reference model has a completely imbalanced preference---meaning 99\% against 1\% or 1\% against 99\%---the algorithm bias of the KL RLHF could turn into preference collapse. To explain this, note that the conditional binary probability distribution between two responses, $y_1$ and $y_2$, for a prompt $x$ takes the form 
\begin{equation*}
\label{thm:pc}
p_\textnormal{rlhf}(y_i)=\frac{p_{\textnormal{ref}}(y_i)p_\textnormal{reward}(y_i)}{p_{\textnormal{ref}}(y_1)p_\textnormal{reward}(y_1)+p_{\textnormal{ref}}(y_2)p_\textnormal{reward}(y_2)}=\begin{cases}
    0 & \textnormal{if}\ \  p_{\textnormal{ref}}(y_i)=0,\\
    1 & \textnormal{if}\ \  p_{\textnormal{ref}}(y_i)=1,\\
\end{cases}\quad i=1,2.
\end{equation*}
This extreme case is also shown in Figure \ref{fig:bias}. The phenomenon of preference collapse, if were true, would be detrimental to decision-making, as it would completely let the LLM discard minority opinions. Our experiments on Llama-2-7B and OPT-1.3B indeed show this phenomenon occurs in practice. For each prompt $x$ in the dataset, we randomly generate two responses, $y_1$ and $y_2$. We then calculate the conditional probabilities of $y_1$ with respect to $y_1$ and $y_2$. As shown in Figure \ref{fig:procedure}(a), it becomes evident that the probabilities $p_{\textnormal{ref}}(y_1)$ are very likely to be close to 0 or 1. In contrast, the probabilities induced by the reward model are more uniform, with probabilities between 0 and 1 for a large number of prompts, as illustrated in Figure \ref{fig:procedure}(b). This phenomenon can be observed in the experiment on OPT-1.3B, as demonstrated in the appendix. 

While the exact cause of this phenomenon is unclear to us at the moment, some evidence shows that it might be attributed to varying lengths of responses \citep{park2024disentangling,chen2024odin}. Recall that LLMs generate each token following the next-token prediction. The probability of a response is in the form of the product of probabilities and the number of terms is the length of the response. A direct consequence is that the probability of a response in general decays exponentially with its length. Thus, for two responses that are generated independently from a prompt, the ratio of their probability is likely to be either very large or small, which is especially the case when the two responses differ significantly in length. 

The algorithmic bias and preference collapse stems from the use of a reference model in the RL objective \eqref{eq:kl_rlhf} and, in particular, continues to occur even if the KL divergence is replaced by any $f$-divergence. The discussion of RLHF under $f$-divergence regularization is provided in the appendix.

To conclude this section, we summarize the key differences between PM RLHF and KL (or general $f$-divergence) RLHF. Both the reward model and the reference model implicitly encode preferences---the former aligned with human preferences, and the latter not. KL RLHF blends the preferences of these two models, which can lead to preference collapse. In contrast, PM RLHF focuses on extracting the preference encoded in the reward model while eliminating the influence of the reference model, thereby preserving human preferences.